\documentclass[10pt,journal,compsoc]{IEEEtran}

\ifCLASSOPTIONcompsoc
  \usepackage[nocompress]{cite}
\else
  \usepackage{cite}
\fi

\ifCLASSINFOpdf
\else
\fi

\usepackage{ragged2e}
\usepackage{amsmath}
\usepackage{algorithm}
\usepackage{algorithmic}
\usepackage{bm}
\usepackage{amsfonts}
\usepackage{booktabs}
\usepackage{multirow}
\usepackage{graphicx}
\usepackage{url}
\usepackage{hyperref}
\hypersetup{hidelinks,
	colorlinks=true,
	allcolors=black,
	pdfstartview=Fit,
	breaklinks=true}

\usepackage{color}
\definecolor{Red}{cmyk}{0,1,1,0}
\definecolor{Green}{cmyk}{1,0,1,0}
\definecolor{Cyan}{cmyk}{1,0,0,0}
\definecolor{Purple}{cmyk}{0.45,0.86,0,0}
\definecolor{Sky}{cmyk}{1.0,0.8,0,0}
\definecolor{Rosolic}{cmyk}{0.00,1.00,0.50,0}
\definecolor{Blue}{cmyk}{1.00,1.00,0.00,0}
\definecolor{Orange}{cmyk}{0,0.52,0.80,0}
\definecolor{Black}{cmyk}{1,0,0,1}

\newcommand{\revise}[1]{{\color{Black}            {#1}}}

\begin{document}

\title{Indoor Scene Reconstruction with Fine-Grained Details Using Hybrid Representation and Normal Prior Enhancement}

\author{
  Sheng Ye,
  Yubin Hu,
  Matthieu Lin,
  Yu-Hui Wen*,
  Wang Zhao,
  Yong-Jin Liu*,\\
  ~\IEEEmembership{Senior Member,~IEEE},
  and Wenping Wang,~\IEEEmembership{Fellow,~IEEE}

\IEEEcompsocitemizethanks{
\IEEEcompsocthanksitem S. Ye, Y.B. Hu, M. Lin, W. Zhao, and Y.-J. Liu are with BNRist, the Department of Computer Science and Technology, Tsinghua University. E-mail: \{yec22, huyb20, yh-lin21, zhao-w19\}@mails.tsinghua.edu.cn, liuyongjin@tsinghua.edu.cn.
\IEEEcompsocthanksitem Y.-H. Wen is with Beijing Key Laboratory of Traffic Data Analysis and Mining, School of Computer and Information Technology, Beijing Jiaotong University. E-mail: yhwen1@bjtu.edu.cn.
\IEEEcompsocthanksitem W.P. Wang is with the Department of Computer Science \& Engineering, Texas A\&M University. E-mail: wenping@tamu.edu.
\IEEEcompsocthanksitem * Corresponding author
}%
}

\markboth{SUBMITTED TO IEEE TVCG}%
{Ye \MakeLowercase{\textit{et al.}}: Indoor Scene Reconstruction with Fine-Grained Details Using Hybrid Representation and Normal Prior Enhancement}

\IEEEtitleabstractindextext{%
\begin{justify}
\begin{abstract}
  The reconstruction of indoor scenes from multi-view RGB images is challenging due to the coexistence of flat and texture-less regions alongside delicate and fine-grained regions. Recent methods leverage neural radiance fields aided by predicted surface normal priors to recover the scene geometry. These methods excel in producing complete and smooth results for floor and wall areas. However, they struggle to capture complex surfaces with high-frequency structures due to the inadequate neural representation and the inaccurately predicted normal priors.
  This work aims to reconstruct high-fidelity surfaces with fine-grained details by addressing the above limitations.
  To improve the capacity of the implicit representation, we propose a hybrid architecture to represent low-frequency and high-frequency regions separately. To enhance the normal priors, we introduce a simple yet effective image sharpening and denoising technique, coupled with a network that estimates the pixel-wise uncertainty of the predicted surface normal vectors. Identifying such uncertainty can prevent our model from being misled by unreliable surface normal supervisions that hinder the accurate reconstruction of intricate geometries.
  Experiments on the benchmark datasets show that our method outperforms existing methods in terms of reconstruction quality. Furthermore, the proposed method also generalizes well to real-world indoor scenarios captured by our hand-held mobile phones. Our code is publicly available at: \href{https://github.com/yec22/Fine-Grained-Indoor-Recon}{https://github.com/yec22/Fine-Grained-Indoor-Recon}.
\end{abstract}
\end{justify}

}

\maketitle

\IEEEdisplaynontitleabstractindextext

\IEEEpeerreviewmaketitle

\IEEEraisesectionheading{\section{Introduction}\label{sec:introduction}}

\IEEEPARstart{3}{D} reconstruction of a target scene from a sequence of multi-view RGB images is a fundamental problem in computer vision, which has a wide range of applications in various fields, including robotics, filming, gaming, virtual reality, and so on.
Specifically, high-fidelity reconstruction of indoor scenes is extremely challenging because indoor scenes have both large, flat areas (floor, wall, roof, \textit{etc}.) and high-frequency, fine-grained areas (small stuff on the table, delicate furniture, \textit{etc}.).

Encoding scenes with a neural implicit function \cite{deepsdf, occupancy, siren} has attracted increasing attention due to its compactness and good performance.
Some recent works have achieved remarkable results in 3D scene reconstruction from 2D supervision, by combining the neural implicit function and differentiable volume rendering. 
NeuS \cite{neus} and VolSDF \cite{volsdf} can reconstruct high-quality watertight surfaces of a single object by representing the geometry as an implicit signed distance field (SDF). Follow-up works try to integrate the implicit SDF representation and additional geometric priors, including semantic segmentation \cite{manhattansdf}, monocular depth \cite{monosdf}, or surface normals \cite{neuris}, to address the challenge of indoor scene reconstruction. Compared to traditional MVS-based \cite{patchstereo, colmap} or TSDF-based \cite{atlas, vortx} approaches, these neural implicit methods can produce complete and promising surfaces, but still struggle with delicate and complex structures.

\begin{figure}[tb]
  \centering 
  \includegraphics[width=\columnwidth]{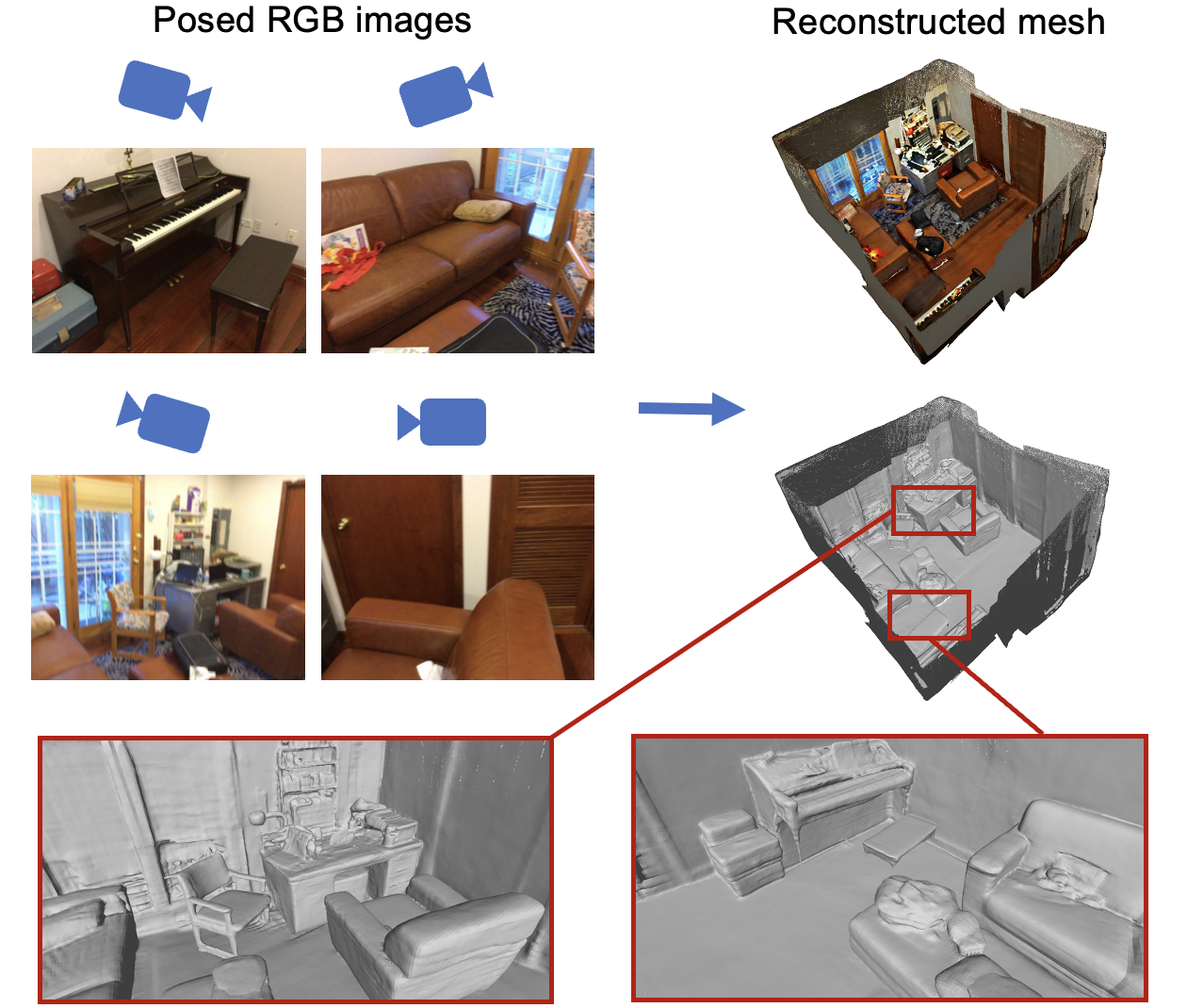}
  \caption{%
    Our proposed method can reconstruct fine and accurate indoor scenes only from a sequence of posed RGB images.
  }
  \label{fig:intro_demo}
  \vspace{-0.1in}
\end{figure}

In this paper, we aim to reconstruct high-fidelity indoor scene surfaces with fine and accurate details utilizing the implicit SDF framework \cite{neus}. We first analyze and conclude two main limitations of the existing neural implicit methods that lead to the failure of fine-grained reconstruction.
One of the limitations is that existing methods typically use a large MLP network to approximate the underlying geometry function of the entire scene. Indeed, they can produce surprisingly good results for object-level surface reconstruction, but are less capable of recovering more complicated indoor scenes. According to SIREN \cite{siren} and MonoSDF \cite{monosdf}, the deep MLP network possesses an inductive smoothness bias, and is therefore suitable to encode surfaces of flat and low-frequency areas. However, indoor scenes also consist of many high-frequency and fine-grained regions, which are difficult for a single MLP to express. Therefore, a novel architecture with greater expressive capacity is needed.
Another limitation is that the surface normal priors used to regularize the implicit SDF may be inaccurate, especially in regions with complex structures. These priors are typically monocular estimated by off-the-shelf neural networks \cite{surface_normal_uncertainty, omnidata}, which tend to produce noisy and erroneous results for thin and delicate structures. Models will be misled to generate wrong geometries when directly supervised by these inaccurate priors. Therefore, it is necessary to improve the quality of the predicted normal priors and design a mechanism to measure their reliability.

To improve the capacity of expressing both the flat areas and fine details, we design a novel hybrid representation. We implicitly decouple an indoor scene into different regions and exploit suitable neural representations to encode these regions separately.
Specifically, we utilize a branch of MLP network to encode the rough outline of the scene, and another branch of tri-plane features with a shallow decoder to represent the fine-grained details.
To recover fine geometric details, some previous studies \cite{go-surf, nice-slam, voxurf} attempt to utilize the voxel grid features. Nevertheless, they require complex training schemes and regularizations since the voxel representation tends to produce noisy surfaces. Moreover, the voxel grids are also memory-intensive due to their cubically growing consumption.
Inspired by the remarkable performance of tri-plane representation in encoding human faces \cite{eg3d, rodin}, we choose this memory-efficient representation instead of voxel grids to encode the delicate details. In our experiments, we empirically find that the tri-plane branch is suitable for expressing high-frequency details, which complements the MLP branch. In addition, our proposed hybrid architecture does not require complicated strategies for training.

As for the enhancement of predicted normal priors, we observe that artifacts present in the input RGB images can greatly affect the quality of predicted normals. Thus, we devise a sharpening and denoising technique before feeding the images into the prior estimation module, which effectively reduces the errors in predicted normals.
Furthermore, we design an uncertainty module, which predicts the pixel-wise uncertainty maps of the estimated normal priors. Besides the RGB images and monocular estimated normals, this module also takes features from the visual foundation model \cite{dino} as input to supplement the prediction with high-level structural information. We then treat the predicted uncertainty as the weight of normal prior supervision. This uncertainty module determines the reliability of normal priors and reduces the negative impact of incorrectly predicted priors on reconstruction quality.
In conclusion, we summarize our contributions as follows:

\begin{itemize}
    \item We propose a novel hybrid implicit SDF architecture that incorporates MLP and tri-plane to better represent the low-frequency and high-frequency regions of indoor scenes simultaneously.
    \item We design an image enhancement technique and an uncertainty module to improve the quality of predicted normal priors and guide our network to effectively leverage more accurate priors. 
    \item Qualitative and quantitative experiments show that results produced by our method are better than state-of-the-art methods.
    Apart from the commonly used datasets, our method also generalizes well to real-world indoor scenes captured by ourselves, demonstrating the potential for practical applications.
\end{itemize}

\begin{figure*}[tb]
  \centering 
  \includegraphics[width=\linewidth]{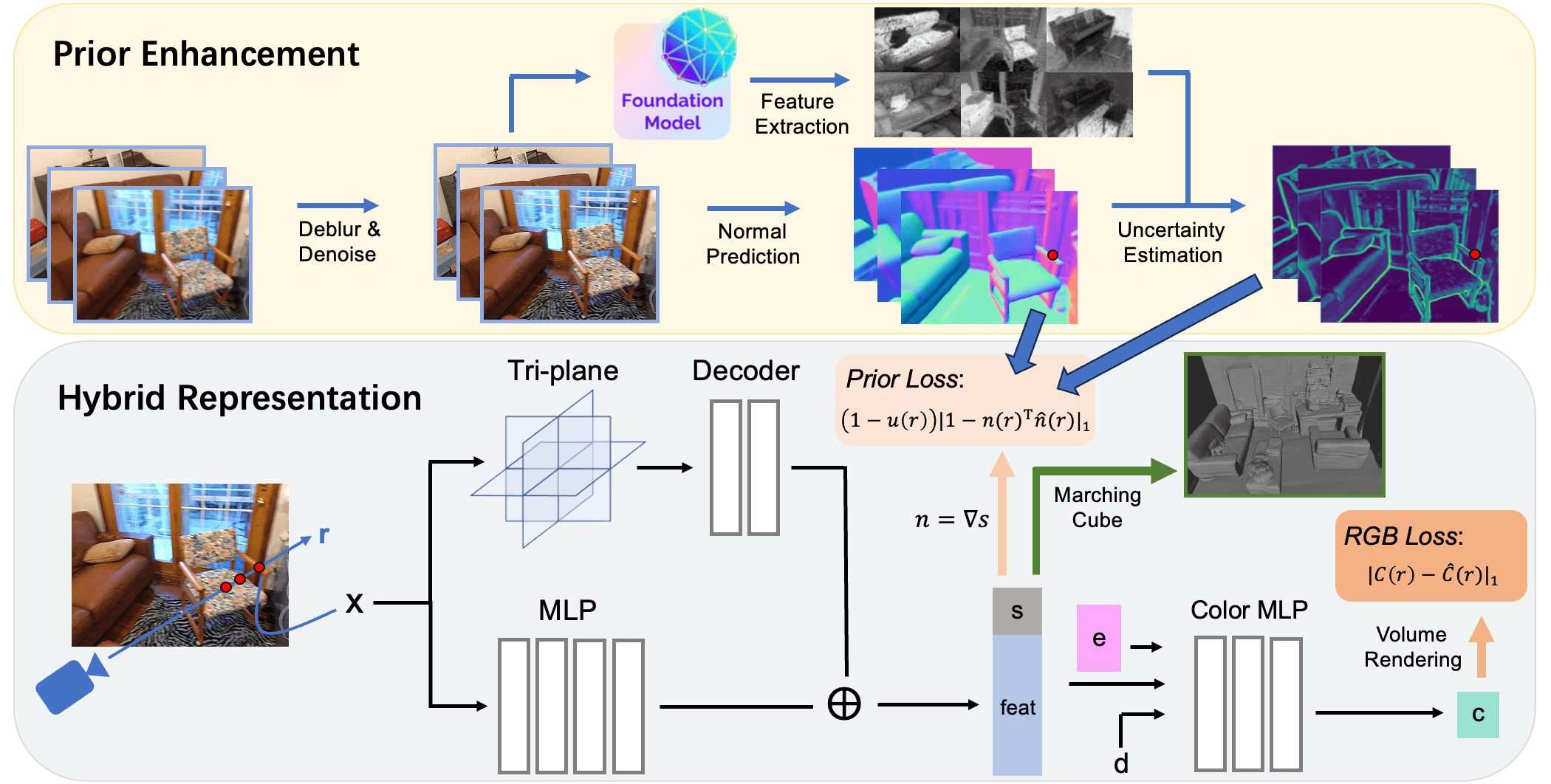}
  \caption{%
    The overall pipeline of our proposed approach. We tackle high-frequency regions in scene reconstruction from the perspective of both the representation and normal priors. In particular, we propose a hybrid geometry representation to enhance the expressive power, and an image preprocessing technique along with pixel-wise uncertainty to enhance the normal priors.
  }
  \label{fig:pipeline}
\end{figure*}

\section{Related Works}

\subsection{Neural Surface Representation}
Representing geometric surfaces by neural implicit functions has recently received increasing attention, because of its compactness and remarkable performance.
The seminal work of DeepSDF \cite{deepsdf} first proposes to use a neural network to model a signed distance field, which encodes the underlying geometry of the target object. However, DeepSDF requires ground-truth 3D meshes to supervise the learning process.
Recent methods incorporate neural implicit functions and differentiable rendering to reconstruct surfaces only from the supervision of multi-view 2D images. DVR~\cite{dvr} first proposes a differentiable rendering formulation for implicit shape and texture representations. IDR \cite{idr} further designs an architecture that simultaneously learns the implicit SDF field, camera poses, and a neural renderer. UNISURF \cite{unisurf} utilizes occupancy to represent implicit surfaces and modifies the volume rendering equation correspondingly. Subsequently, NeuS \cite{neus} and VolSDF \cite{volsdf} define the density used in volume rendering process as logistic sigmoid function and Laplace’s cumulative distribution function applied to an SDF representation.
Nevertheless, all those works use a single MLP to approximate the implicit function, which struggles to scale to large and complicated scenes due to the network capacity. There also exist other researches \cite{nice-slam, go-surf, dvgo, instant-ngp} that use multi-scale voxel grids followed by a shallow decoder to represent the scenes, but this leads to non-smoothness. Neuralangelo~\cite{neuralangelo} leverages 3D hash grids to recover dense 3D surface structures. However, hash grids also suffer from localities and hash collisions. In this work, we devise a novel hybrid representation incorporating MLP and tri-plane to enhance the network's ability to express high-frequency and delicate surfaces, while maintaining non-local smoothness.

\subsection{Indoor Scene Reconstruction}
Conventional multi-view stereo (MVS) methods \cite{colmap, patchstereo} often struggle to reconstruct large texture-less indoor scene areas densely. With the development of large-scale indoor datasets~\cite{scannet, replica},
learning-based MVS methods are proposed to alleviate this problem. These methods~\cite{mvsnet, dpsnet, simplerecon} usually predict a depth map for each frame and fuse these depth maps to build the final scene. However, the resulting surfaces tend to be noisy and incomplete due to the depth inconsistency problem. Another research branch~\cite{atlas, neuralrecon, vortx, semanticfusion} proposes directly regressing input images to truncated signed distance function (TSDF) volume. Then, the meshes can be extracted from TSDF using the marching-cube algorithm. Due to the limitation of TSDF volume resolution, the generated surfaces often lack details.

Inspired by NeRF's \cite{nerf} excellent performance in novel view synthesis tasks, some recent works attempt to leverage implicit surface representation and 
differentiable volume rendering to reconstruct indoor scenes. Considering the complexity of indoor scenes, additional priors are also provided to recover a plausible geometry. ManhattanSDF \cite{manhattansdf} utilizes extra 2D semantic segmentation to detect wall and floor regions, and applies geometry regularization based on Manhattan-world assumption. NeuRIS \cite{neuris} and MonoSDF~\cite{monosdf} exploit monocular estimated normal and depth priors to reconstruct smooth surfaces. NeuRIS also devises a novel cross-view geometric checking technique. HelixSurf \cite{helixsurf} attempts to combine PatchMatch-based MVS and neural radiance fields, and proposes a joint optimization pipeline. Although these neural implicit methods significantly improve the reconstruction quality compared to conventional methods, surfaces in complex and fine-grained regions are still unsatisfactory. Furthermore, few studies have explored how to reduce the negative impact of inaccurate priors on the final reconstruction results.

\subsection{Uncertainty Estimation}
Uncertainty estimation, which aims to quantify the reliability of predictions, can help recognize failure scenarios and enable robust applications.
Uncertainty is crucial for many computer vision tasks such as optical flow~\cite{ilg2018uncertainty}, SLAM~\cite{kerl2013dense}, and multi-view stereo~\cite{wang2022itermvs}. Predicted uncertainty is used to facilitate the optimization through robust loss~\cite{teed2021droid}, guided sampling~\cite{sinha2019variational}, outlier rejection~\cite{raguram2009exploiting}, \textit{etc}.
However, the integration of the neural radiance field with uncertainty remains a relatively unexplored terrain. Some current efforts \cite{s-nerf, cf-nerf} mainly focus on the task of rendering, while we focus on surface reconstruction. In this work, we novelly leverage the estimated uncertainty in a weighted loss to guide our model to utilize more precise priors while avoiding misguidance by less accurate ones.

\begin{figure*}[t]
  \centering 
  \includegraphics[width=\linewidth]{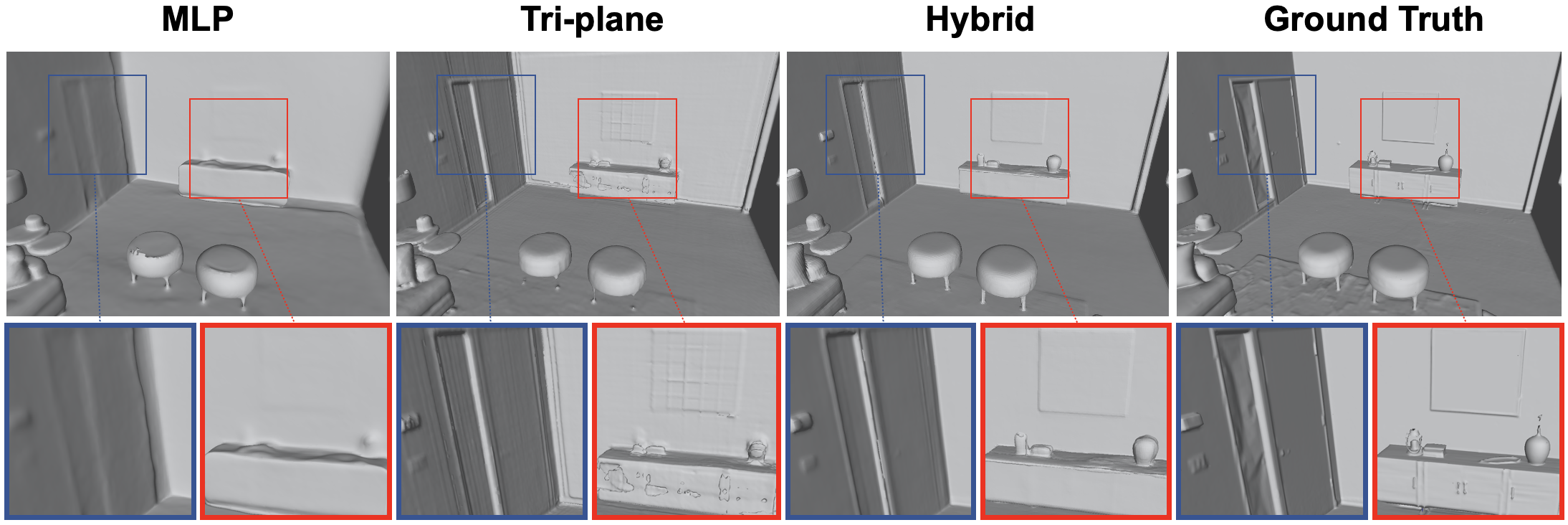}
  \caption{%
    Comparison of the reconstructed meshes generated by different representations. Zoom in for better visualization.
    %
  }
  \label{fig:arch_comp}
\end{figure*}

\section{Method}

In this work, our goal is to reconstruct high-fidelity, room-scale surfaces with fine and accurate details from multiple calibrated RGB images. We represent the scene geometry as an implicit signed distance field encoded by our proposed hybrid architecture.
Our architecture can learn 3D geometric information only from 2D image supervision by applying SDF-based volume rendering. As indoor scenes often contain flat and textureless areas, we also utilize estimated surface normal priors as additional geometric constraints. To effectively leverage normal priors, we design an enhancement technique and an uncertainty mechanism for better reconstruction quality. Figure \ref{fig:pipeline} summarizes the overall pipeline of our proposed approach.

In the subsequent sections, we first briefly revisit the preliminary of the SDF-based volume rendering equation. And then, we discuss and describe the details of our proposed hybrid geometric representation, prior enhancement techniques, and uncertainty estimation module. Finally, we introduce the losses that are used to optimize our model.

\subsection{Preliminary}
A continuous 3D scene can be modeled as a signed distance field $f_g$ and a color field $f_c$~\cite{neus, neuris}. 
Both of the two fields are typically represented by neural networks. The differentiable volume rendering technique can learn the signed distance and color fields only from 2D image supervision.
We denote a ray emitted from a viewing camera as $\bm{r}(t) = \bm{o} + t\bm{d}$, where $\bm{o}$ is the camera center and $\bm{d}$ is the viewing direction. Given a set of sample points $\{ \bm{r}(t_i) | i=1,...,N \}$ along the ray, the corresponding pixel color of this ray can be calculated as

\begin{equation}
\label{eqn:volume_render}
C(\bm{r}) = \sum_{i=1}^{N}{T_i\alpha_i f_c(\bm{r}(t_i),\bm{d})}, \ T_i = \prod_{j=1}^{i-1}{(1 - \alpha_j)},
\end{equation}
where $T_i$ is the accumulated transmittance, and $\alpha_i$ denotes the opacity of the $i$-th ray segment and is defined as

\begin{equation}
\alpha_i=\max \left(0, \frac{\Phi_\tau(f_g(\bm{r}(t_{i}))) - \Phi_\tau(f_g(\bm{r}(t_{i+1})))}{\Phi_\tau(f_g(\bm{r}(t_{i})))} \right),
\end{equation}
where $\Phi_\tau$ is the Sigmoid function with a learnable parameter $\tau$.
Then, the difference between the rendered pixel color $C(\bm{r})$ and the ground-truth pixel color $\hat{C}(\bm{r})$ is minimized to optimize the $f_g$ and $f_c$. After optimization, the surface $\mathcal{S}$ can be extracted as the zero level-set of the signed distance field $\mathcal{S} = \{\bm{x}|f_g(\bm{x})=0\}$.

\subsection{Hybrid Representation for Geometry}
\label{sec:hybrid_architecture}

Previous methods \cite{manhattansdf, neuris, helixsurf} usually utilize a single MLP to represent the geometry of the entire indoor scene. Although equipped with positional encodings that encompass high-frequency bands, these methods still tend to produce low-frequency surfaces and struggle to reconstruct complex and delicate structures. Recent researches \cite{siren, monosdf} point out that the reasons may lie in the smoothness bias of MLP networks.
Studies \cite{monosdf, nice-slam} have also tried to replace the MLP with voxel grid features to recover fine details. However, the voxel representation is less compact due to its cubic memory consumption and is prone to generating noisy and inconsistent surfaces, which hinders the reconstruction of flat areas that frequently appear in indoor settings.

To encode the high-frequency features of indoor scenes effectively and efficiently, we attempt to use the tri-plane representation. As optimization updates only propagate to local plane features instead of all parameters (\textit{i.e.}, locality), we find that tri-plane representations are better at expressing fine-grained details than MLP.
Besides, tri-planes are also memory-efficient, as their memory consumption only grows quadratically with the resolution. As shown in Figure \ref{fig:arch_comp}, our preliminary experiments indicate the potential of using the tri-plane for scene reconstruction. The reconstructed 3D meshes of the tri-plane representation are sharper and possess more details than those generated by MLP representation. However, due to the locality nature of the tri-plane, we also observe that using the tri-plane representation alone will cause unwanted artifacts on planar regions.

Empirically, we find that different areas of indoor scenes have different characteristics. The outer contours of a scene are usually smooth and flat, while the interior often contains delicate structures. Thus, in order to accurately reconstruct both the flat areas and the fine-grained details, we propose a hybrid geometry architecture, which contains an MLP branch and a tri-plane branch (see Figure \ref{fig:pipeline}). We aim to exploit the smoothness bias of MLP to encode the rough outline of the scene and the high-frequency bias of tri-plane features to encode the delicate structures.
To encourage the two branches to focus on different areas of the scene, we find that the initialization plays a vital role. \revise{Following SAL \cite{sal}, we initialize the MLP branch to produce approximate SDF of a unit bounding sphere, and initialize the tri-plane branch to produce a zero SDF.} In this way, we implicitly decouple an indoor scene into two distinct regions without the need for extra segmentation annotations.
\revise{Given a 3D position $\bm{x} \in \mathbb{R}^3$, our pipeline passes it through the MLP branch to get a coarse SDF $\tilde{\bm{s}}$ and a coarse hidden feature $\tilde{\bm{h}}$. Simultaneously, the tri-plane branch outputs the high-frequency complements.}
Our tri-plane representation $T$ consists of three axis-aligned orthogonal feature planes. Each feature plane has the dimension of $N \times N \times C$, where $N$ is the spatial resolution, and $C$ is the number of channels. We retrieve the final feature $[T_{xy}, T_{xz}, T_{yz}]$ by projecting $\bm{x}$ onto these planes to find the tri-plane features and concatenating them together. A shallow, two-layer decoder further regresses the aggregated feature $[T_{xy}, T_{xz}, T_{yz}]$ to the residual SDF $\Delta \bm{s}$ and residual hidden feature $\Delta \bm{h}$. The final outputs are the summation of these two branches

\begin{equation}
\bm{s} = \tilde{\bm{s}} + \Delta \bm{s}, \ \bm{h} = \tilde{\bm{h}} + \Delta \bm{h}.
\end{equation}

Subsequently, we feed the hidden feature $\bm{h}$ and the viewing direction $\bm{d}$ into the color network to obtain the view-dependent emitted radiance $\bm{c}$. Inspired by NeRF-W~\cite{nerf-w}, we also attach a learnable appearance embedding $\bm{e}$ for each view as an extra input to the color network. Appearance embeddings aim to compensate for photometric and environmental variations between images and guarantee a multi-view consistent reconstruction. Our experiments demonstrate that cleaner and sharper surfaces can be obtained by using appearance embeddings.

During the training process, we optimize the hybrid geometry architecture, color network, and appearance embeddings simultaneously.
As shown in Figure \ref{fig:arch_comp}, our hybrid architecture combines the advantages of both MLP and tri-plane, preserving the sharp details and eliminating the artifacts as well.
In conclusion, our method leverages characteristics of different regions of indoor scenes and, for the first time, exploits suitable neural representations to encode these regions separately.

\subsection{Prior Enhancement}

\subsubsection{Image Sharpening and Denoising Techniques}
Surface normal priors can serve as globally consistent geometric constraints to improve the reconstruction quality. Thus, we use the monocular predicted normal priors as extra supervision, which can regularize the SDF especially in textureless regions where color supervision is insufficient. Specifically, we can obtain the surface normal $\bm{n}$ under certain viewpoints using a similar volume accumulation process as Equation (\ref{eqn:volume_render}):

\begin{equation}
\bm{n}(\bm{r}) = \sum_{i=1}^{N}{T_i\alpha_i\nabla\bm{s}_i},
\end{equation}

where $\nabla\bm{s}_i$ is the gradient of SDF at different sample points. During training, we minimize the difference between the normal calculated by our model and the pseudo ground-truth normal estimated by off-the-shelf tools \cite{surface_normal_uncertainty, omnidata}.

It is worth noting that the estimated normal priors are not always accurate. We observe that artifacts presented in the input RGB images, especially noise and motion blur, can significantly affect the quality of estimated normal priors. These artifacts are prevalent in real-world scenarios. Thus, we propose to use a sharpening and denoising image enhancement before predicting the normal maps. For the sharpening process, we use a $3 \times 3$ kernel to perform convolution on the whole input image. This kernel can be regarded as an edge detector, which amplifies the difference between the center pixel and its surrounding pixels. Formally, this operation is defined as 

\begin{equation}
p_i' = 9 \cdot p_i - \sum_{p_j \in \mathcal{N}_{i}}{p_j},
\end{equation}

where $p_i$ is the center pixel and $\mathcal{N}_{i}$ is the eight neighboring pixels of $p_i$ inside the $3 \times 3$ kernel. The sharpening process alleviates motion blur and strengthens edges in the images, facilitating the subsequent normal estimation of thin structures. For the denoising process, we employ the median filtering technique that replaces an image pixel with the median value of its neighbors. This operation can effectively remove the isolated noise while maintaining the sharpness of the image. We find that error regions of the estimated normal priors are reduced after applying the denoising operation (more details can be found in our experiments).
Although we apply the sharpening and denoising per image, the appearance embeddings introduced in Section \ref{sec:hybrid_architecture} can handle the slight cross-view inconsistencies caused by image enhancement.
Our sharpening and denoising techniques indicate that image processing in 2D can indeed affect the reconstruction quality in 3D, and the bridge between this is surface normal and volume rendering.

\subsubsection{Uncertainty Estimation Module}
Our sharpening and denoising techniques can reduce the normal estimation errors caused by imperfect image quality. However, it is still difficult to predict accurate surface normal maps of complex and intricate regions, even with high-quality input images. 
Directly regularizing the implicit SDF field with these inaccurate surface normal predictions can result in degenerated geometries that lack delicate structures.
Therefore, we attach a pixel-wise uncertainty map to each estimated normal map, which reflects the reliability of different regions of the normal maps.

For regions with lower uncertainty values, our method depends more on the estimated surface normal priors to achieve fast and accurate geometry reconstruction.
For delicate regions with higher uncertainty values, our method assigns a smaller weight to the surface normal loss, such that the optimization of the SDF field depends more on the realistic RGB information and avoids being misled by inaccurate surface normal predictions.
Accordingly, we design an uncertainty weighted normal prior loss function to assign larger (smaller) weights to the low (high) uncertainty areas when using these normal priors for supervision, which can be formulated as follows,

\begin{equation}
\mathcal{L}_{prior} = \frac{1}{| \mathcal{R} |} \sum_{\bm{r} \in \mathcal{R}} {(1 - \bm{u}(\bm{r})) \left \| 1 -  \bm{n}(\bm{r})^\top \hat{\bm{n}}(\bm{r}) \right \|_1},
\end{equation}

where $\mathcal{R}$ is the set of rays in each batch. $\bm{n}(\bm{r})$ and $\hat{\bm{n}}(\bm{r})$ are the rendered normal and the pseudo ground-truth normal.
Specifically, for the uncertainty value corresponding to a ray $\bm{r}$, we normalize it to $[0, 1]$, denoted as $\bm{u}(\bm{r})$. 
When $\mathcal{L}_{prior}$ supervises the implicit SDF field together with the typical color loss, our proposed uncertainty values can be viewed as a trade-off between the color information and normal information. 
For complicated regions with high uncertainty, the proposed loss reduces the weight of normal supervision, thereby strengthening the role of color supervision.
By balancing the supervision weights, our model can resist being misled by unreliable normal priors and effectively utilize more reliable normal priors.

\begin{figure}[t]
  \centering 
  \includegraphics[width=\columnwidth]{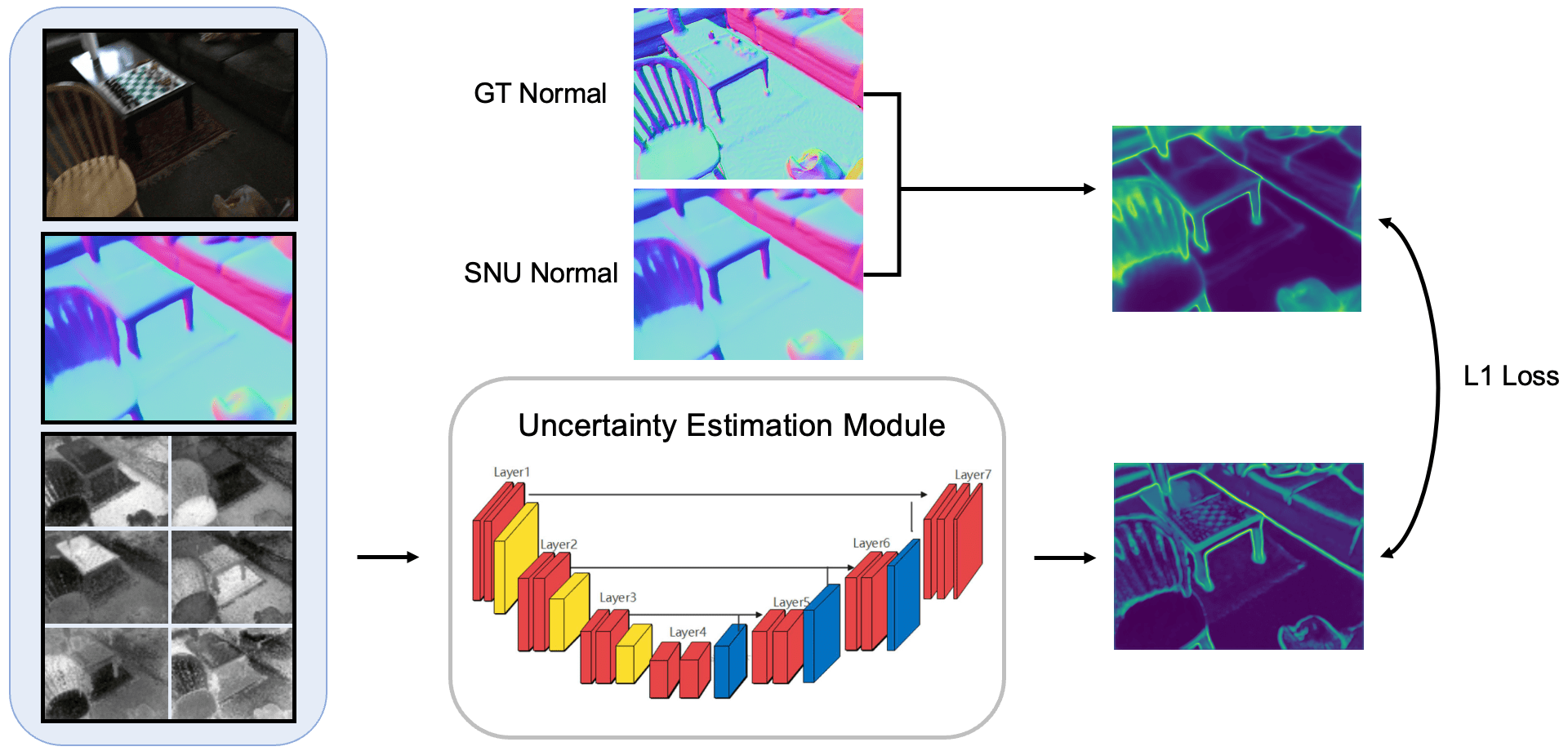}
  \caption{%
    We propose an uncertainty estimation module to predict the pixel-wise uncertainty maps of the normal priors.
  }
  \label{fig:uncertainty}
\end{figure}

To predict the pixel-wise uncertainty of estimated normal maps, we design a novel module shown in Figure \ref{fig:uncertainty}. This module adopts a U-Net \cite{unet} architecture with RGB images and the estimated normal maps as input. 
\revise{To facilitate the uncertainty prediction, we also input high-level DINO \cite{dino} features, since the normal uncertainty is usually structure-relevant.
We choose the visual foundation model DINO rather than CLIP \cite{clip}, as DINO tends to capture pixel-level structure-aware image features during its self-supervised learning process, while CLIP features tend to be global and linguistic-related.
Furthermore, a previous work~\cite{VFM} also demonstrates the effectiveness of DINO features in facilitating downstream visual understanding tasks, outperforming other foundation models (\textit{e.g.}, CLIP \cite{clip} and DeiT \cite{deit}).} We extract features from the last attention layer of DINO, and then apply PCA technique to retain the most essential information.
We train this uncertainty estimation module on a subset (disjoint from the test set) of the ScanNet \cite{scannet}. Concretely, we compute the difference between the estimated normals and normals rendered from real 3D meshes to obtain the ground-truth uncertainty maps, and use them to supervise our proposed estimation module.

\begin{table}[t]
\centering
\caption{Definitions of the metrics to evaluate the 3D reconstruction quality.}
\setlength{\tabcolsep}{12pt}
\renewcommand\arraystretch{1.5}
\begin{tabular}{cc}
\toprule
3D Metric & Definition \\
\midrule
\textit{Accuracy} & mean$_{p \in P}$(min$_{p^* \in P^*}$ $\| p - p^* \|$) \\
\textit{Completeness} & mean$_{p^* \in P^*}$(min$_{p \in P}$ $\| p - p^* \|$) \\
\textit{Precision} & mean$_{p \in P}$(min$_{p^* \in P^*}$ $\| p - p^* \|$ $<$ \textit{threshold}) \\
\textit{Recall} & mean$_{p^* \in P^*}$(min$_{p \in P}$ $\| p - p^* \|$ $<$ \textit{threshold}) \\
\textit{F-Score} & $\frac{2 \times Precision \times Recall}{Precision + Recall}$ \\
\bottomrule
\end{tabular}
\label{table:metric_def}
\end{table}

\subsection{Loss Functions}

The overall loss is the weighted sum of normal prior loss $\mathcal{L}_{prior}$, Eikonal loss $\mathcal{L}_{eik}$, and RGB color loss $\mathcal{L}_{rgb}$:

\begin{equation}
\mathcal{L} = \lambda_p \mathcal{L}_{prior} + \lambda_{e} \mathcal{L}_{eik} + \lambda_r \mathcal{L}_{rgb}.
\end{equation}

All reasonable signed distance fields must satisfy the Eikonal equation \cite{eikonal}, which constrains the gradient of SDF to be equal to 1. Thus, the Eikonal loss $\mathcal{L}_{eik}$ to regularize the gradient of SDF is denoted by

\begin{equation}
\mathcal{L}_{eik} = \frac{1}{N} \sum_{i=1}^{N} {(\left \| \nabla\bm{s}_i \right \|_2 - 1)^2},
\end{equation}

where $N$ is the total number of sampled points, and $\nabla\bm{s}_i$ is the gradient of SDF at different sample points. The RGB color loss $\mathcal{L}_{rgb}$ is defined as

\begin{equation}
\mathcal{L}_{rgb} = \frac{1}{| \mathcal{R} |} \sum_{\bm{r} \in \mathcal{R}} {\left \| C(\bm{r}) - \hat{C}(\bm{r}) \right \|_1},
\end{equation}

where $C(\bm{r})$ and $\hat{C}(\bm{r})$ are the predicted and ground-truth RGB colors for ray $\bm{r}$ respectively, and $\mathcal{R}$ is the set of rays in each batch.

\begin{table}[t]
\centering
\caption{Quantitative comparison of reconstruction quality on ScanNet dataset (threshold = 0.05). Bold indicates the best.}
\setlength{\tabcolsep}{6.0pt}
\begin{tabular}{c|ccccc}
\toprule
Methods & Acc$\downarrow$ & Comp$\downarrow$ & Prec$\uparrow$ & Recall$\uparrow$ & F-score$\uparrow$ \\
\midrule
COLMAP & 0.047 & 0.235 & 0.711 & 0.441 & 0.537 \\
\midrule
NeuS & 0.179 & 0.208 & 0.313 & 0.275 & 0.291 \\
NeuralAngelo & 0.132 & 0.109 & 0.505 & 0.467 & 0.485 \\
\midrule
ManhattanSDF & 0.072 & 0.068 & 0.621 & 0.586 & 0.602 \\
NeuRIS & 0.051 & 0.048 & 0.720 & 0.674 & 0.696 \\
MonoSDF & 0.035 & 0.048 & 0.799 & 0.681 & 0.733 \\
HelixSurf & 0.038 & 0.044 & 0.786 & 0.727 & 0.755 \\
\midrule
Ours & \textbf{0.033} & \textbf{0.041} & \textbf{0.814} & \textbf{0.737} & \textbf{0.773} \\
\bottomrule
\end{tabular}
\label{table:scannet_comparison}
\end{table}

\begin{table}[t]
\centering
\caption{Quantitative comparison of reconstruction quality on Replica dataset (threshold = 0.05). Bold indicates the best.}
\setlength{\tabcolsep}{6.0pt}
\begin{tabular}{c|ccccc}
\toprule
Methods & Acc$\downarrow$ & Comp$\downarrow$ & Prec$\uparrow$ & Recall$\uparrow$ & F-score$\uparrow$ \\
\midrule
COLMAP & 0.030 & 0.095 & 0.872 & 0.530 & 0.658 \\
\midrule
NeuS & 0.187 & 0.290 & 0.201 & 0.133 & 0.161 \\
NeuralAngelo & 0.103 & 0.125 & 0.445 & 0.392 & 0.417 \\
\midrule
ManhattanSDF & 0.145 & 0.194 & 0.611 & 0.427 & 0.501 \\
NeuRIS & 0.034 & 0.070 & 0.803 & 0.712 & 0.754  \\
MonoSDF & 0.024 & 0.030 & 0.933 & 0.898 & 0.915 \\
\midrule
Ours & \textbf{0.022} & \textbf{0.026} & \textbf{0.942} & \textbf{0.931} & \textbf{0.936} \\
\bottomrule
\end{tabular}
\label{table:replica_comparison}
\end{table}

\section{Experiments}

\subsection{Experiment Details} 

\begin{figure*}[tb]
  \centering 
  \includegraphics[width=\linewidth]{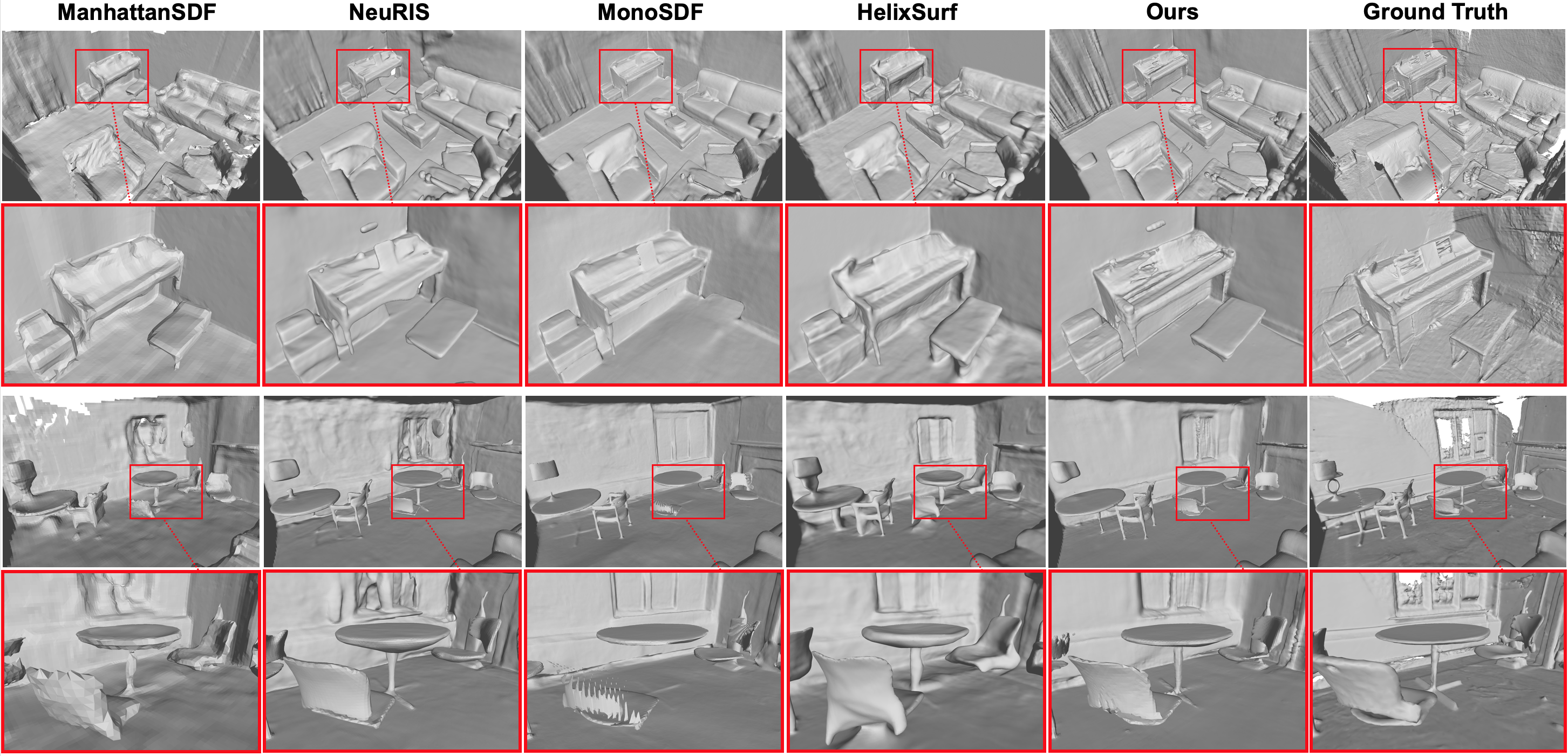}
  \caption{%
    Qualitative comparison of reconstructed meshes with other baselines on ScanNet dataset.
  }
  \label{fig:vis_comp_scannet}
\end{figure*}

\begin{figure*}[tb]
  \centering 
  \includegraphics[width=\linewidth]{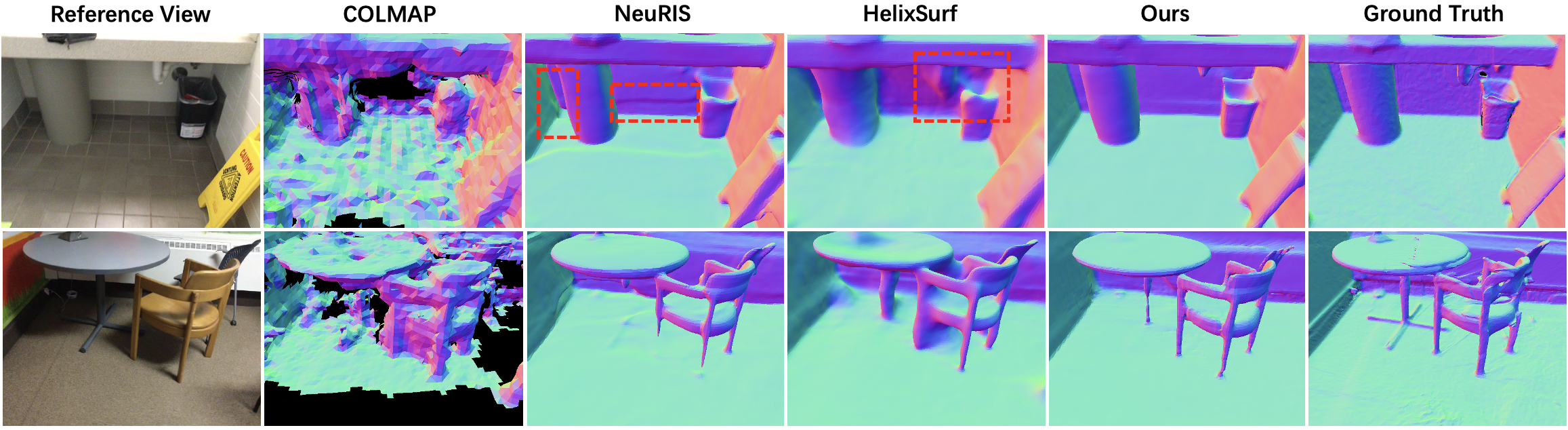}
  \caption{%
    Qualitative comparison of rendered normal maps with other baselines on ScanNet dataset.
  }
  \label{fig:normal_comp_scannet}
\end{figure*}

\subsubsection{Datasets}
We mainly conduct the quantitative and qualitative experiments on two commonly used benchmark datasets: ScanNet \cite{scannet} and Replica \cite{replica}. Furthermore, we also demonstrate the generalization ability of our method on several real-world indoor scenes captured by ourselves.

\noindent\textbf{ScanNet.} It is a real-world indoor dataset containing various scenes captured with Kinect V1 RGB-D cameras. The BundleFusion \cite{bundlefusion} is then used to provide camera poses and 3D meshes. For ScanNet, we use the same test split from ManhattanSDF \cite{manhattansdf}. 

\noindent \textbf{Replica.} It is a synthetic dataset which provides high-quality RGB images, dense geometry, and semantic annotations.
For Replica, we follow the settings of MonoSDF \cite{monosdf}.

\begin{figure*}[tb]
  \centering 
  \includegraphics[width=\linewidth]{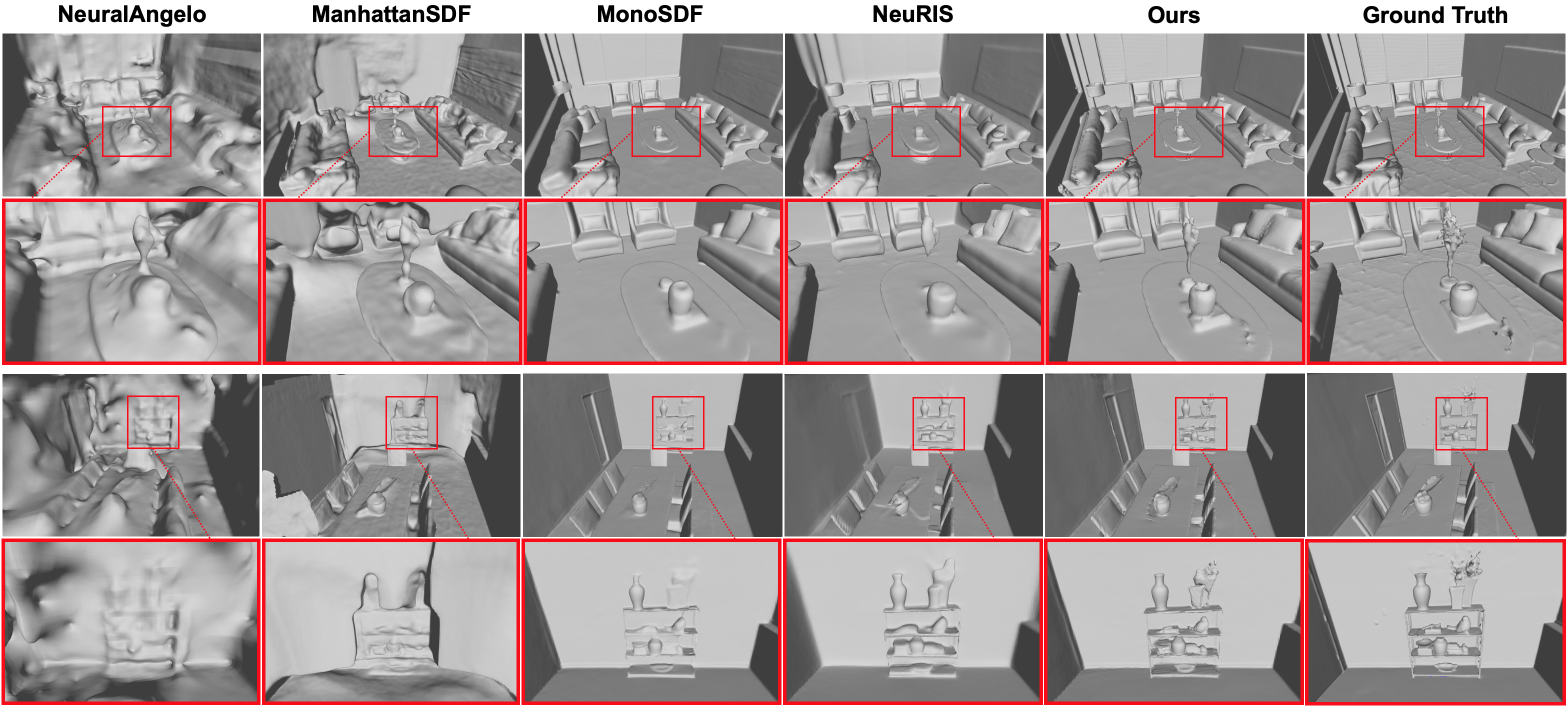}
  \caption{%
    Qualitative comparison of reconstructed meshes with other baselines on Replica dataset.
  }
  \label{fig:vis_comp_replica}
\end{figure*}

\begin{figure*}[tb]
  \centering 
  \includegraphics[width=\linewidth]{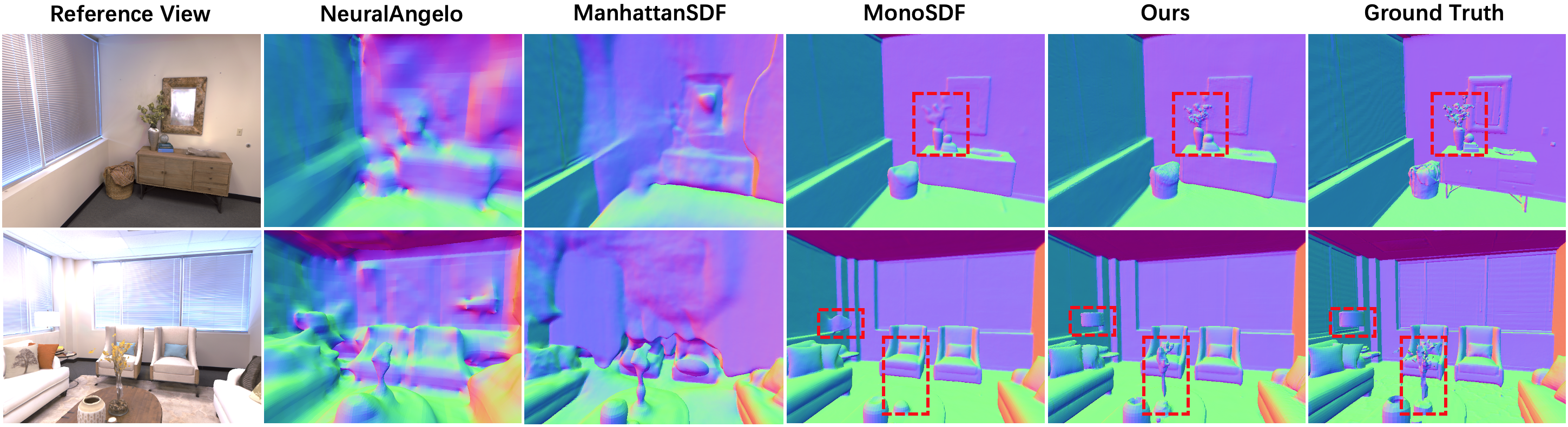}
  \caption{%
    Qualitative comparison of rendered normal maps with other baselines on Replica dataset.
  }
  \label{fig:norm_comp_replica}
\end{figure*}

\subsubsection{Compared Methods}
We compare our method with the following baselines: (1) Traditional MVS method COLMAP \cite{colmap}; (2) State-of-the-art neural implicit surface reconstruction methods, including NeuS \cite{neus} and NeuralAngelo \cite{neuralangelo}; (3) State-of-the-art neural indoor scene reconstruction methods, including ManhattanSDF \cite{manhattansdf}, NeuRIS \cite{neuris}, MonoSDF \cite{monosdf}, and HelixSurf \cite{helixsurf}. Since HelixSurf does not fully open source the data preparation code and provides only the preprocessed data on ScanNet, we exclude HelixSurf from experiments on Replica dataset. All baselines are trained from scratch. 

\subsubsection{Evaluation Metrics}
Following previous methods \cite{atlas, manhattansdf, neuris, helixsurf}, we use five standard metrics to evaluate the reconstruction geometry quality: \textit{Accuracy}, \textit{Completeness}, \textit{Precision}, \textit{Recall}, and \textit{F-score}. These metrics are defined in Table \ref{table:metric_def}, where $P$ and $P^*$ denote the sample points from the predicted and ground-truth mesh.

\subsubsection{Implementation}
Our proposed model is experimented on an NVIDIA A100 GPU. For the geometry network, we adopt an 8-layer MLP, and $256 \times 256 \times 16$ dimension tri-plane features. For the color network, we adopt a 4-layer MLP. The uncertainty estimation module leverages a U-Net architecture with skip connections. Specifically, this uncertainty module contains four downsample convolution blocks and four upsample blocks. We select 300 scenes (disjoint from the test set) from the ScanNet and construct 12696 data pairs to train this module.
We implement our model in Pytorch using Adam optimizer. The loss weights are set to $\lambda_p = \lambda_r = 1.0$, and $\lambda_e = 0.1$. In each batch, $|\mathcal{R}| = 512$. The training process takes 40k - 50k iterations depending on the scene's complexity, and it typically costs 1.5 - 2 hours.

\subsection{Comparisons} 

\subsubsection{Quantitative Evaluation}
Table \ref{table:scannet_comparison} and Table \ref{table:replica_comparison} summarize the quantitative comparison results on ScanNet dataset and Replica dataset. We compare our method with several baselines and report the averaged metrics.
Although NeuS \cite{neus} and NeuralAngelo~\cite{neuralangelo} can reconstruct high-fidelity object-level surfaces, their performance drops drastically in indoor scenarios. This is because the images captured from indoor scenes tend to be less idealized (artifacts are common) and overlap less between images, making the model hard to optimize. Aided by extra priors, methods like ManhattanSDF \cite{manhattansdf}, NeuRIS \cite{neuris}, and MonoSDF \cite{monosdf} achieve performance improvement. However, the \textit{Recall} metrics of these methods are usually low, indicating that the reconstruction results may lack fine-grained details. HelixSurf \cite{helixsurf} attempts to incorporate MVS point clouds to produce more complete surfaces. Nevertheless, the MVS points can be noisy and degrade the \textit{Precision} metric of the reconstruction meshes.

On both datasets, our proposed method surpasses existing methods on all five metrics. \textit{F-score} is usually regarded as a faithful metric for evaluating geometry quality as it considers both accuracy and completeness. In particular, we achieve remarkable improvement on the \textit{F-score} metric, which indicates that our approach can produce high-quality reconstructed meshes with fine and accurate details.

\subsubsection{Qualitative Evaluation}
We visualize the meshes reconstructed by different methods in Figure \ref{fig:vis_comp_scannet} and Figure \ref{fig:vis_comp_replica}. For better comparison, we also render the normal maps of these meshes in Figure \ref{fig:normal_comp_scannet} and Figure \ref{fig:norm_comp_replica}.
Conventional COLMAP \cite{colmap}, which is based on feature matching, struggles to reconstruct large texture-less indoor areas densely.
NeuralAngelo \cite{neuralangelo} fails to generate smooth and consistent surfaces, possibly due to the discontinuity nature of the hash grid.
Despite ManhattanSDF~\cite{manhattansdf} introducing extra constraints based on Manhattan-world assumption, the precision of the reconstructed mesh is still limited.
NeuRIS \cite{neuris} and MonoSDF~\cite{monosdf} leverage geometric priors estimated by off-the-shelf networks and adopt large MLPs to encode the entire scene. Because of the limited expression capacity of MLP and the inaccuracy that existed in monocular estimated priors, NeuRIS tends to produce artifacts in complex and intricate regions. Similarly, MonoSDF cannot faithfully reconstruct thin and fine-grained structures.
HelixSurf can indeed generate more complete meshes, but at the expense of accuracy. As shown in Figure \ref{fig:vis_comp_scannet} and Figure \ref{fig:normal_comp_scannet}, surfaces produced by HelixSurf are rough and blunt in corner and edge areas.
Compared with baselines, our method is able to express both high-frequency details and low-frequency smoothness, and produces clean and visually appealing results.

\subsection{Ablation Study}

\begin{table}[t]
\centering
\caption{Ablation study of different geometry representations.}
\setlength{\tabcolsep}{1.8pt}
\begin{tabular}{c|c|ccccc}
\toprule
Config & Setting & Prec$\uparrow$ & Recall$\uparrow$ & F-score$\uparrow$ & Mem. & Param. \\
\midrule
\multirow{2}{*}{MLP only} &  layer=8 & \multirow{2}{*}{0.773} & \multirow{2}{*}{0.693} & \multirow{2}{*}{0.729} & \multirow{2}{*}{13GB} & \multirow{2}{*}{\textbf{2.3M}} \\
& hidden=512 \\
\midrule
\multirow{2}{*}{Grids only} & reso=256 & \multirow{2}{*}{0.751} & \multirow{2}{*}{0.686} & \multirow{2}{*}{0.716} & \multirow{2}{*}{24GB} & \multirow{2}{*}{537M} \\
& channel=32 \\
\midrule
\multirow{2}{*}{Tri-plane only} & reso=512 & \multirow{2}{*}{0.782} & \multirow{2}{*}{0.722} & \multirow{2}{*}{0.750} & \multirow{2}{*}{\textbf{9GB}} & \multirow{2}{*}{25.7M} \\
& channel=32 \\
\midrule
\multirow{3}{*}{MLP + Grids} & reso=256 & \multirow{3}{*}{0.783} & \multirow{3}{*}{0.711} & \multirow{3}{*}{0.745} & \multirow{3}{*}{18GB} & \multirow{3}{*}{269M} \\
& channel=16 \\
& hidden=256 \\
\midrule
\multirow{3}{*}{MLP + Tri-plane} & reso=256 & \multirow{3}{*}{\textbf{0.814}} & \multirow{3}{*}{\textbf{0.737}} & \multirow{3}{*}{\textbf{0.773}} & \multirow{3}{*}{11GB} & \multirow{3}{*}{4.5M} \\
& channel=16 \\
& hidden=256 \\
\bottomrule
\end{tabular}
\label{table:ablation_representation}
\end{table}

\begin{figure}[t]
  \centering 
  \includegraphics[width=\columnwidth]{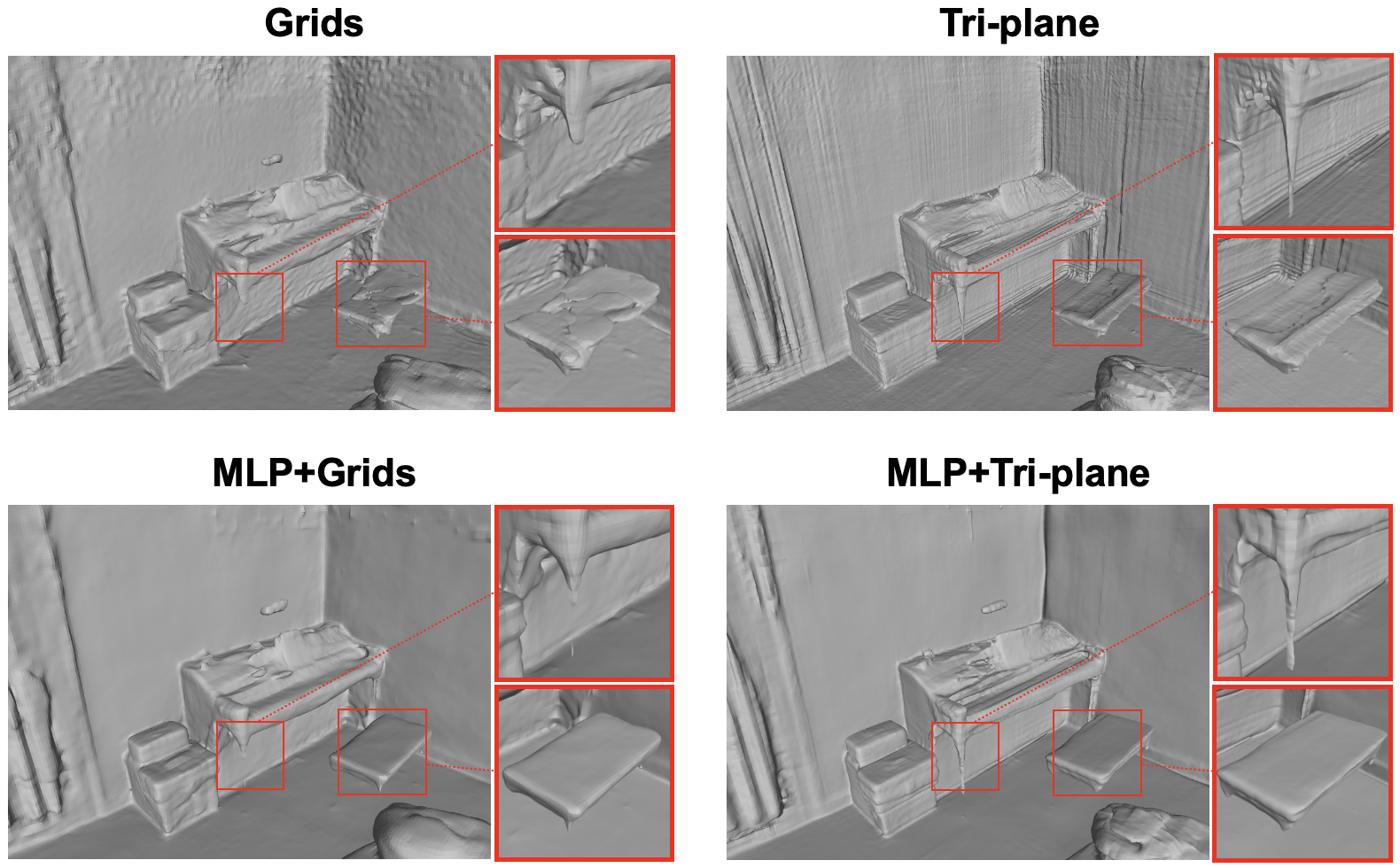}
  \caption{%
    Comparison of reconstructed meshes produced by different geometry representations.
  }
  \label{fig:arch_ablation}
\end{figure}

We conduct ablation experiments on ScanNet by separately changing or removing one of these components in our proposed method: (a) the hybrid geometry representation; (b) the appearance embeddings (w/o embed); (c) the prior enhancement (w/o enhance); (d) the uncertainty estimation module (w/o uncertainty, photometric uncertainty). Table \ref{table:ablation_representation} and Table \ref{table:ablation_others} show the results of our ablation study.

\begin{figure}[t]
  \centering 
  \includegraphics[width=\columnwidth]{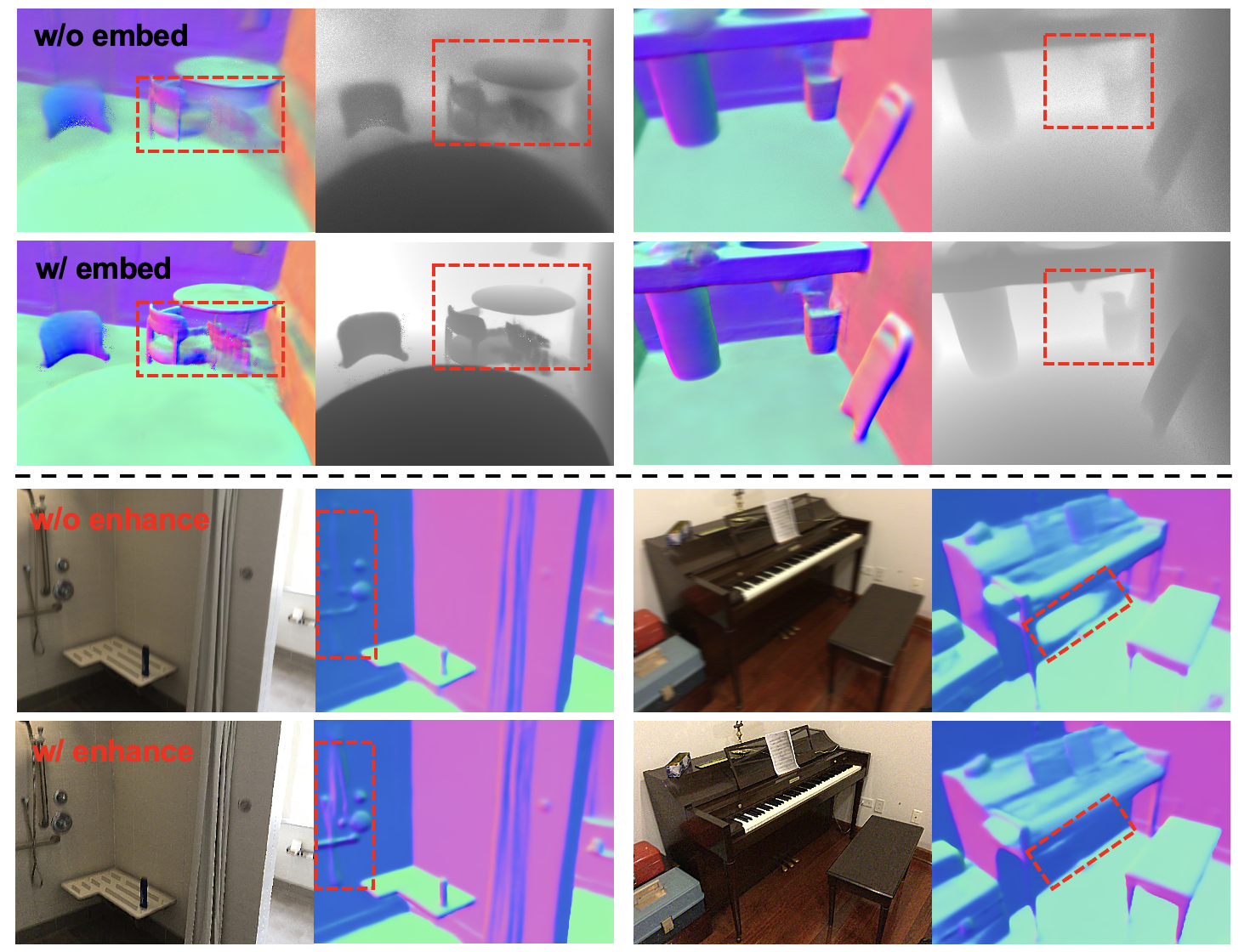}
  \caption{%
    We visualize the effects of appearance embeddings (top) and prior enhancement (bottom). Zoom in for better comparison.
  }
  \label{fig:ablation}
\end{figure}

\begin{figure}[t]
  \centering 
  \includegraphics[width=\columnwidth]{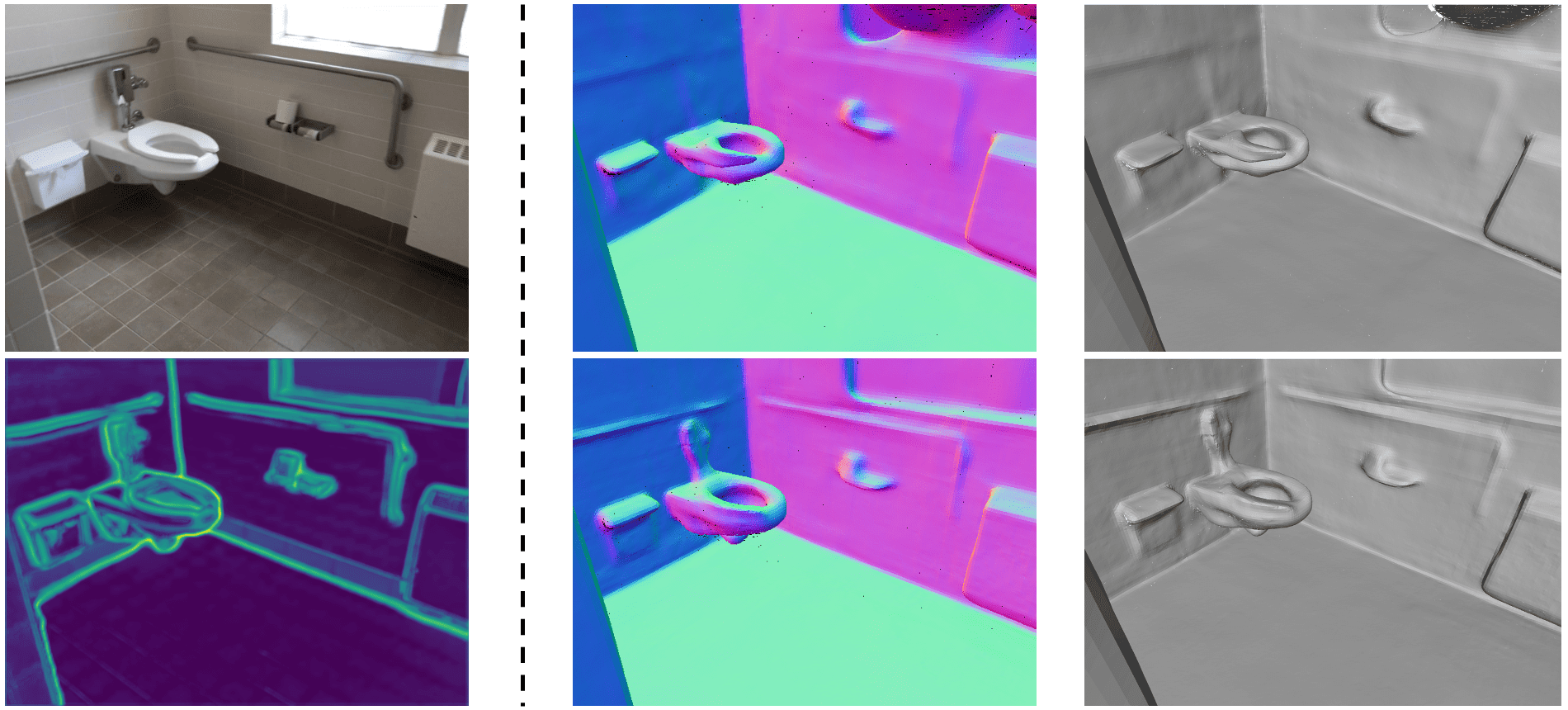}
  \caption{%
   We visualize the reconstruction results with / without the uncertainty estimation (bottom / top).
  }
  \label{fig:detail}
\end{figure}

\begin{table}[b]
\centering
\caption{Ablation study of different components of our approach.}
\setlength{\tabcolsep}{11.0pt}
\begin{tabular}{c|ccccc}
\toprule
Config & Prec$\uparrow$ & Recall$\uparrow$ & F-score$\uparrow$ \\
\midrule
w/o embed & 0.798 & 0.721 & 0.757 \\
w/o enhance & 0.774 & 0.719 & 0.745 \\
w/o uncertainty & 0.808 & 0.728 & 0.766 \\
photometric uncertainty & 0.810 & 0.729 & 0.768 \\
\midrule
full model & \textbf{0.814} & \textbf{0.737} & \textbf{0.773}\\
\bottomrule
\end{tabular}
\label{table:ablation_others}
\end{table}

\subsubsection{Analysis of Geometry Representation}

\begin{figure*}[t]
  \centering 
  \includegraphics[width=\linewidth]{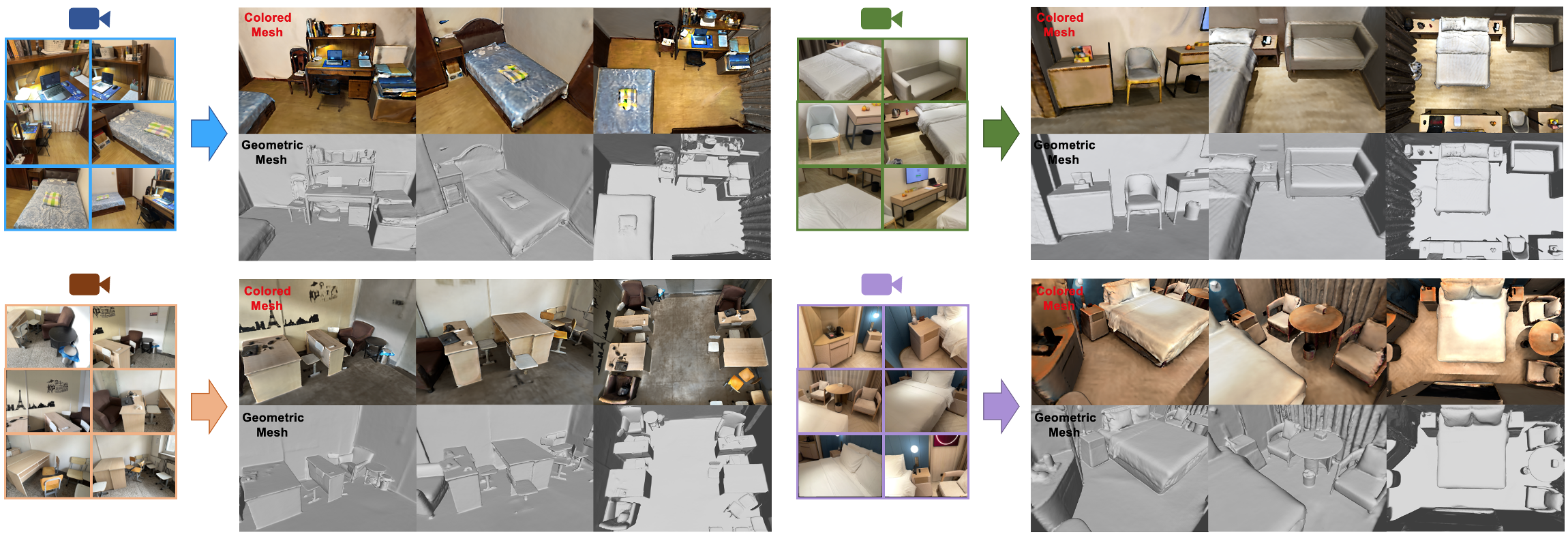}
  \caption{%
   Our method generalizes well to real-world indoor scenarios captured by hand-held mobile phones.
  }
  \label{fig:real_world}
\end{figure*}

\begin{figure*}[t]
  \centering 
  \includegraphics[width=\linewidth]{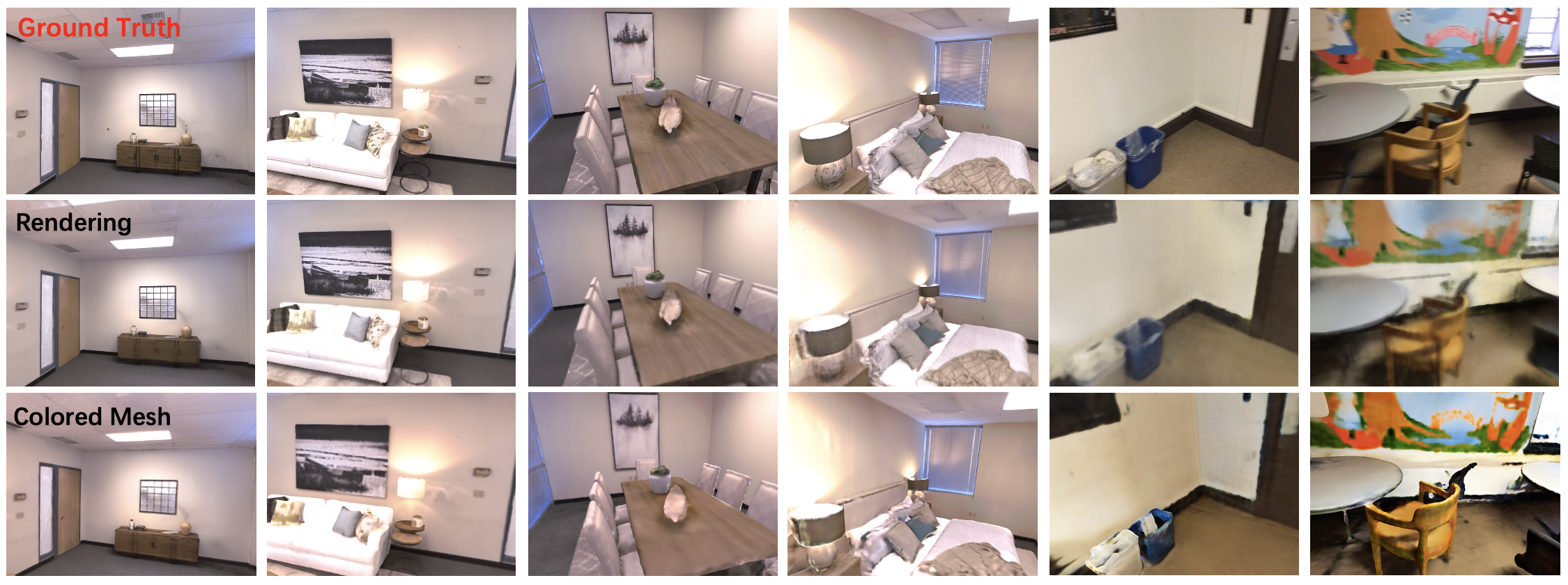}
  \caption{%
    \revise{Results of the novel view synthesis task. The first row is the ground-truth images, the second and third rows are novel views generated by volume rendering and rasterized by our reconstructed colored mesh, respectively.}
  }
  \label{fig:novel_view}
\end{figure*}

In Table \ref{table:ablation_representation}, we compare the effectiveness of several geometry representations and report their parameters and memory consumption. 
MLP is a parameter compact representation but tends to generate over smooth surfaces, as discussed in Section \ref{sec:hybrid_architecture}. Although voxel grids can encode high-frequency details \cite{monosdf}, they produce noisy and bumpy surfaces in planar regions (shown in Figure \ref{fig:arch_ablation}), which lead to the degradation of overall metrics. Also, the memory consumption and parameters of voxel grids increase significantly. In contrast, tri-plane is memory efficient and enables higher resolution to encode intricate structures. Metric improvements also indicate that tri-plane is a powerful geometry representation. Nevertheless, some axis-aligned striped artifacts still exist.

We find that using hybrid architecture (MLP+Grids, MLP+Tri-plane) can consistently achieve better results than using voxel grids or tri-plane alone, because it combines the advantage of both representations.
Due to the superior expressive power of tri-plane, the hybrid representation of MLP and tri-plane surpasses the hybrid representation of MLP and voxel grids.
Figure \ref{fig:arch_ablation} also shows that MLP+Triplane is better at encoding thin and delicate structures than MLP+Grids, indicating the geometry representation we adopt is optimal.
Note that for hybrid architecture, we halve the hidden units of the MLP branch and feature channels of the grids/tri-plane branch, as we assume the learning burden of each branch is reduced. Therefore, performance improvements mainly come from the architectural design rather than the model size.

\subsubsection{Analysis of Appearance Embedding}
Since we only supervise our model on 2D image observations, the inconsistency caused by photometric and environmental variations can lead to noisy geometries. Our experiments illustrate that the appearance embeddings can alleviate this problem, as the embeddings can learn to compensate for variations and guarantee a consistent reconstruction. We render the reconstructed normal maps and depth maps in Figure \ref{fig:ablation}. The depth maps and normal maps are sharper and cleaner by adding appearance embeddings.

\subsubsection{Analysis of Prior Enhancement}
As illustrated in Figure \ref{fig:ablation}, imperfections like motion blur during the video capture process can cause degraded predictions of normal priors, which hinder the subsequent reconstruction of fine and accurate geometries. After applying our proposed enhancement, the image becomes sharper, and the predicted normal maps get better. As a result, Table \ref{table:ablation_others} shows that the accuracy and completeness of reconstructed meshes are improved as well.

\subsubsection{Analysis of Uncertainty Estimation}
\begin{figure}[t]
  \centering 
  \includegraphics[width=\columnwidth]{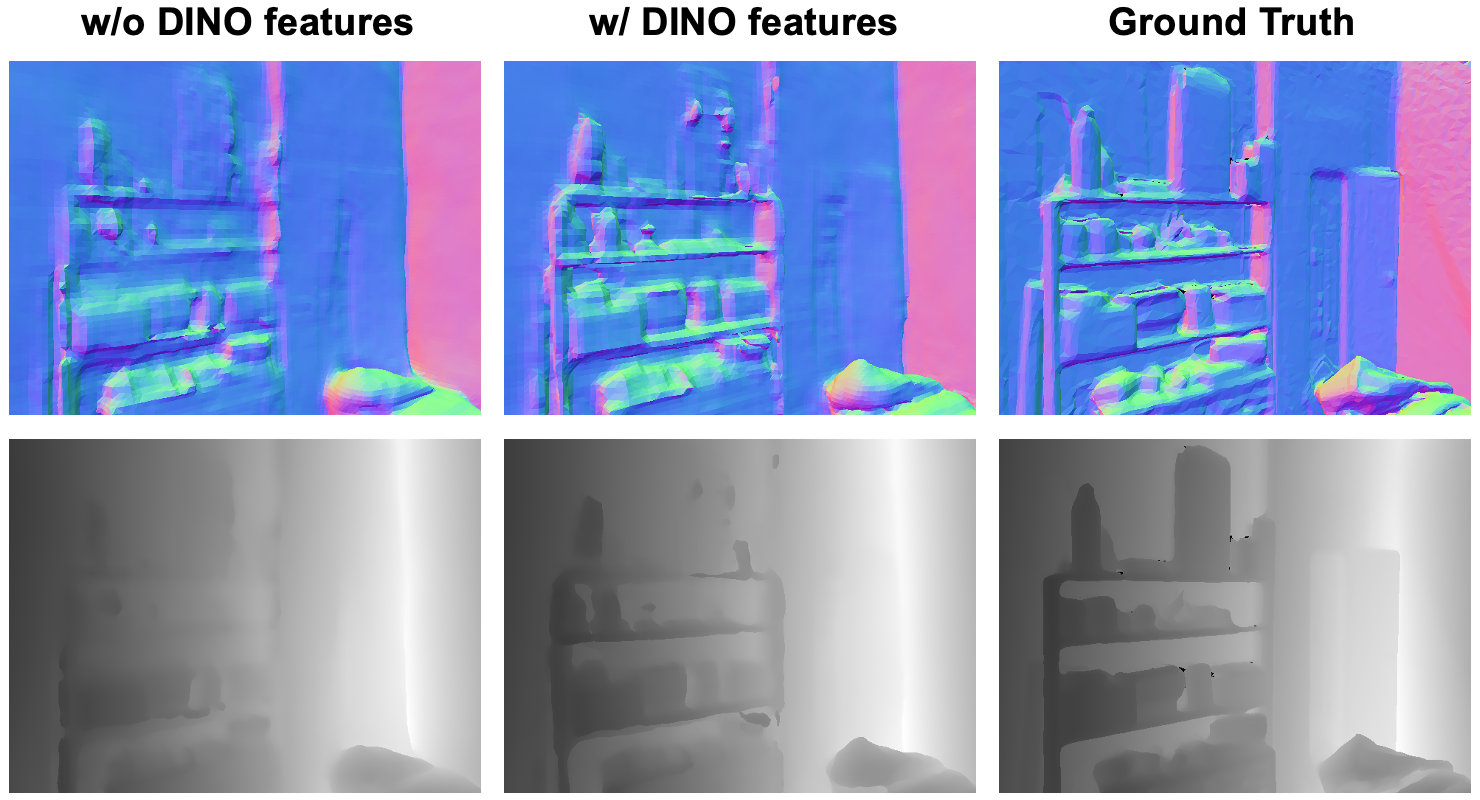}
  \caption{%
    High-level DINO features can further facilitate the reconstruction of intricate details. We visualize the normal and depth maps.
  }
  \label{fig:dino}
\end{figure}

The estimated uncertainty can serve as a measure of reliability and a balance between color supervision and normal supervision. In intricate areas, it guides our network to learn better geometry from rich color information rather than inaccurate normal information, thereby facilitating the correct reconstruction of exquisite structures.
Ablation results show that the uncertainty estimation module helps to improve geometry quality. As fine-grained details take up only  a small proportion of the scene and cannot be fully reflected by numerical metrics, we also visualize the reconstructed meshes and rendered normal maps in Figure \ref{fig:detail}. Adding uncertainty as guidance makes the reconstruction results more visually appealing and richer in high-frequency details. 
In addition, we find that supplementing the uncertainty estimation module with high-level DINO features can further facilitate the reconstruction of complex and delicate structures (shown in Figure \ref{fig:dino}).

\begin{table}[t]
\centering
\caption{Resource consumption compared to several baseline methods.}
\setlength{\tabcolsep}{2.0pt}
\begin{tabular}{c|ccccc}
\toprule
Method & ManhattanSDF & MonoSDF & NeuralAngelo & HelixSurf &  Ours \\
\midrule
Param. & 1.1M & 0.8M & 366M & \textbf{0.2M} & 4.5M  \\
Mem.   & 12GB & 16GB  & 21GB & 20GB  & \textbf{11GB} \\
\revise{FLOPs} & \revise{45.3G} & \revise{36.5G} & \revise{\textbf{34.9G}} & \revise{47.1G} & \revise{44.8G} \\
\bottomrule
\end{tabular}
\label{tab:efficiency}
\end{table}

We have also tried to estimate uncertainty by evaluating the cross-view photometric consistency based on PatchMatch technique, whose performance is worse than using our proposed estimation module (see Table \ref{table:ablation_others}). We assume this is because the artifacts of captured images (jittering or lighting) can largely affect the cross-view photometric consistency, while our estimation module is trained on many data pairs and obtains a certain degree of robustness.

\subsection{Real World Scenarios} 

In order to evaluate the generalization and practicality of our method, we also conduct experiments on real-world indoor scenarios captured by ourselves. Specifically, we walk around each scene and capture a two-minute video with our hand-held mobile phone. We extract 400-500 frames from this video and utilize OpenMVG \cite{openmvg} to estimate camera poses. Then, we use the proposed model to reconstruct 3D surfaces from the calibrated RGB images. Figure \ref{fig:real_world} shows that our approach can produce high-fidelity meshes under different types of indoor scenes, which shows the potential practical value of our method.

\subsection{Novel View Synthesis}

The novel view is another form of understanding 3D scenes apart from meshes. 
\revise{Figure \ref{fig:novel_view} shows the novel views generated by volume rendering, rasterized by our reconstructed colored mesh, and the corresponding ground-truth images.}
Note that while our primary goal is to reconstruct high-fidelity, detailed surfaces of an indoor scenario, we can also synthesize realistic images under novel views facilitated by the accurate reconstruction results.

\subsection{Efficiency and Robustness}

In Table \ref{tab:efficiency}, we compare the resource consumption of several baseline methods. ManhattanSDF \cite{manhattansdf} and MonoSDF \cite{monosdf} adopt the parameter compact MLP models but struggle to reconstruct fine geometries.
The memory consumption and parameters of NeuralAngelo \cite{neuralangelo} are large because it utilizes 3D hash grids, which can be regarded as a special version of voxel grids.
HelixSurf \cite{helixsurf} represents the implicit SDF field with a relatively small MLP accompanied by a PatchMatch-based MVS module. However, the joint optimization of MVS module and neural radiance field in HelixSurf consumes a large memory overhead. \revise{In contrast, our method can produce the best quality meshes with comparable parameters and FLOPs, and less memory consumption.}

\begin{figure}[t]
  \centering 
  \includegraphics[width=\columnwidth]{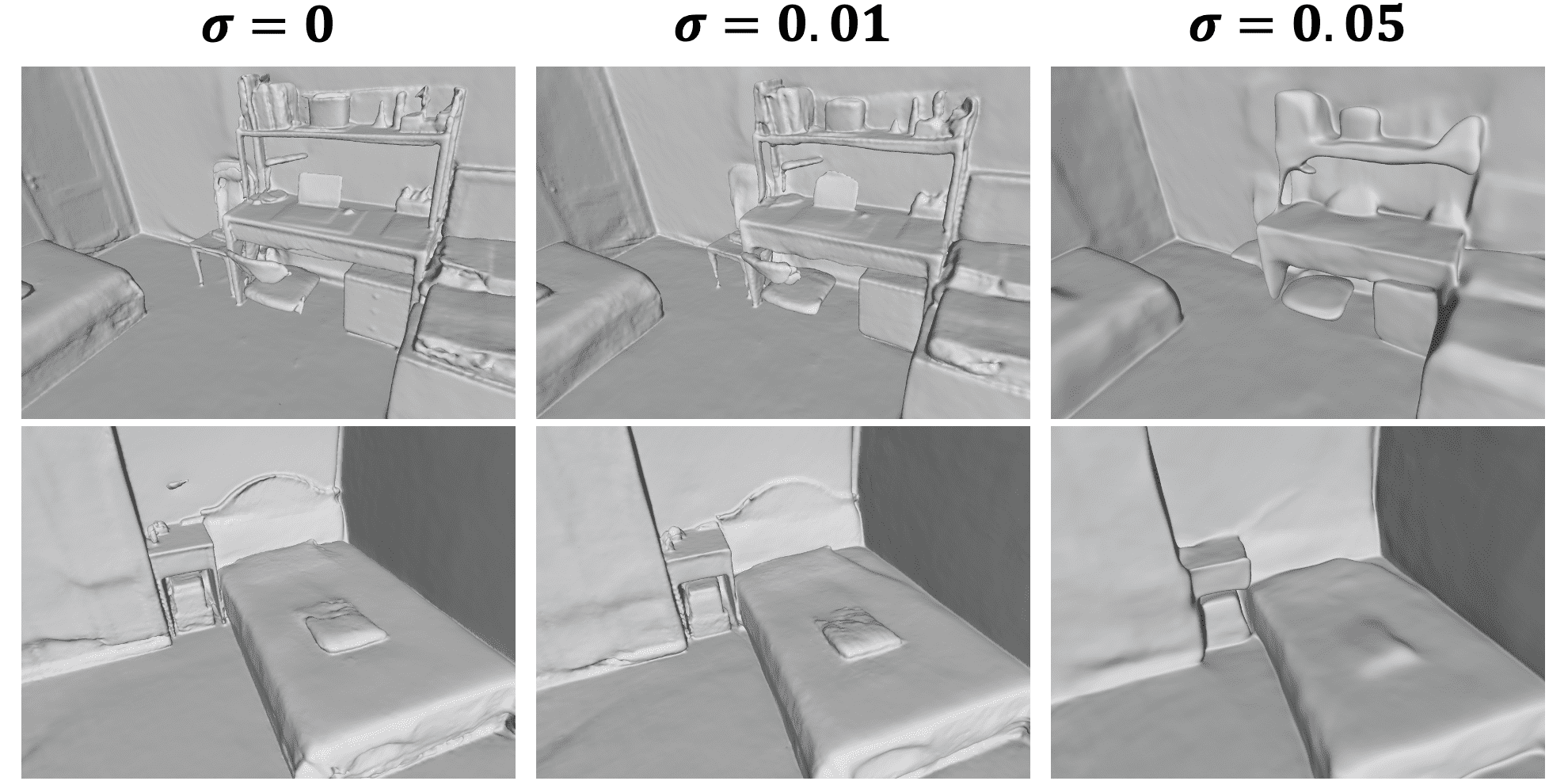}
  \caption{%
    Reconstructed meshes under noisy camera poses ($\sigma$ indicates the noise intensity). Note that our approach can still produce reasonable surfaces even though the camera poses are inaccurate.
  }
  \label{fig:robustness}
\end{figure}

In this work, we assume the camera poses in datasets are precise. However, the camera calibration process may introduce some errors under real circumstances.
To evaluate the robustness of our method under extremely noisy poses, we add a random perturbation (between $[0, \sigma]$) to the camera pose of each view and conduct the 3D reconstruction. 
As illustrated in Figure \ref{fig:robustness}, our method can still generate plausible geometry under noisy camera poses. The robustness is likely due to the uncertainty mechanism and normal priors, which provide the network with valuable cues and regularizations under imperfect poses.

\section{Conclusion and Limitation}

This paper proposes a novel framework to reconstruct indoor scenes with fine-grained details from posed RGB images. To enhance the expressive capability of the network, we design a hybrid geometry representation to encode low-frequency and high-frequency structures separately. This hybrid representation incorporates the advantages of MLP and tri-plane.
Besides, we propose a sharpening and denoising enhancement as well as an uncertainty estimation module. The enhancement technique can facilitate the prediction of sharper and clearer normal priors, and the estimated uncertainty maps can prevent our model from being misled by unreliable priors in intricate regions.
Quantitative and qualitative experiments show that our approach surpasses existing state-of-the-art methods. Our proposed model also generalizes well to real-world scenarios captured by hand-held mobile phones. 

As our method is based on neural radiance fields with volume rendering, the proposed method fails when reconstructing mirrors or glasses. The reason is that the current volume rendering pipeline does not take into account physical reflections and refractions. This can possibly be solved by combining neural radiance fields with Whitted or Monte Carlo ray tracing, which we leave as future work. Furthermore, speeding up the training process for real-time reconstruction and extending our method to very large scenes are also interesting research directions.

\section*{Acknowledgments}
This work was supported by Natural Science Foundation of China (62332019, U2336214, 62202257), and The Talent Fund of Beijing Jiaotong University (2023XKRC045).

\bibliographystyle{IEEEtran}
\bibliography{journal}

.
\ifCLASSOPTIONcaptionsoff
  \newpage
\fi


\begin{IEEEbiography}[{\includegraphics[width=1in,height=1.25in,clip,keepaspectratio]{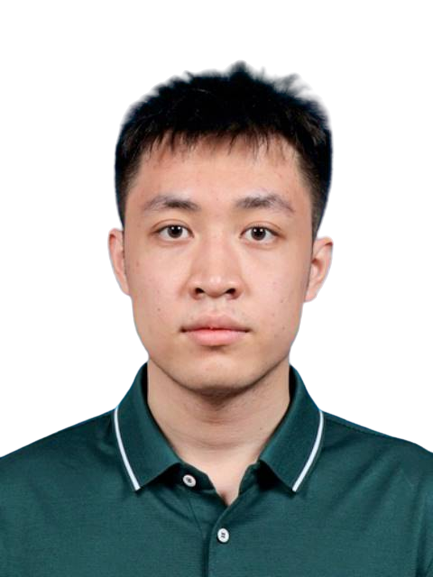}}]
{Sheng Ye} is a Ph.D. student with the Department of Computer Science and Technology, Tsinghua University, China. He received his BEng degree from Tsinghua University, China in 2022. His research interests include computer vision and 3D reconstruction.
\end{IEEEbiography}

\begin{IEEEbiography}[{\includegraphics[width=1in,height=1.25in,clip,keepaspectratio]{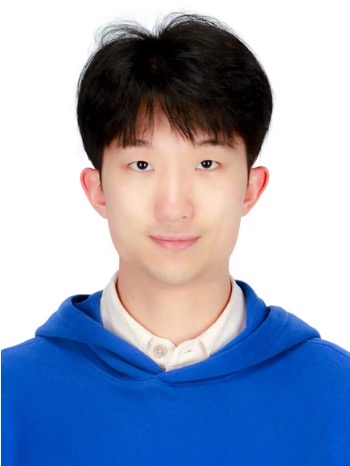}}]{Yubin Hu} is a Ph.D. student with the Department of Computer Science and Technology, Tsinghua University, China. He received his BEng degree from the Electronic Engineering Department, Tsinghua University, China in 2020. He was a Visiting Undergraduate Research Intern of the Harvard John A. Paulson School of Engineering and Applied Sciences in 2019. His research interests include computer vision and 3D reconstruction.
\end{IEEEbiography}

\begin{IEEEbiography}[{\includegraphics[width=1in,height=1.25in,clip,keepaspectratio]{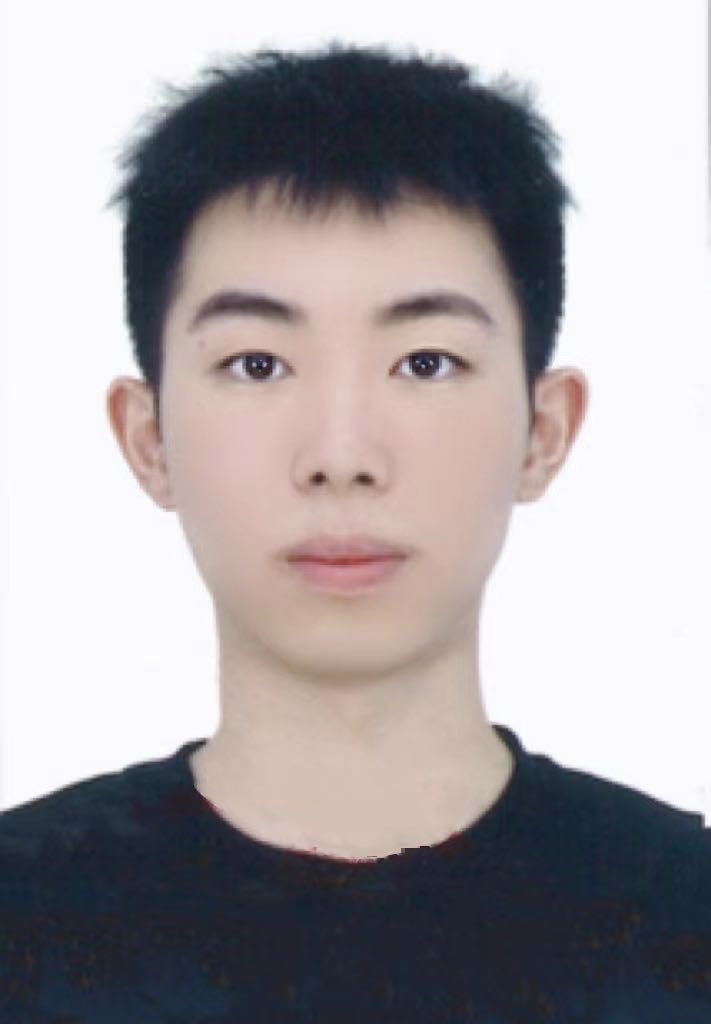}}]
{Matthieu Lin} is a Ph.D. student with the Department of Computer Science and Technology at Tsinghua University, under the supervision of Professor Yong-Jin Liu. He received his B.S.E degree in Computer Science from ESIEA Paris in 2018 and his M.S. degree in Computer Science from Tsinghua University in 2021. His research interests include reinforcement learning and computer vision.
\end{IEEEbiography}

\begin{IEEEbiography}[{\includegraphics[width=1in,height=1.25in,clip,keepaspectratio]{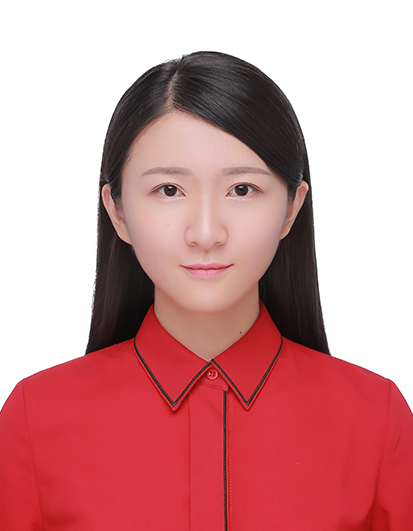}}]
{Yu-Hui Wen} is an associate professor with the Beijing Key Laboratory of Traffic Data Analysis and Mining, School of Computer and Information Technology, Beijing Jiaotong University, Beijing, China. She received her bachelor's degree from Harbin Institute of Technology (HIT), and the Ph.D. degree in computer science and technology from University of Chinese Academy of Sciences (UCAS), Beijing, China, in 2020. Her research interests include machine vision, computer graphics and human motion analysis.
\end{IEEEbiography}

\begin{IEEEbiography}[{\includegraphics[width=1in,height=1.25in,clip,keepaspectratio]{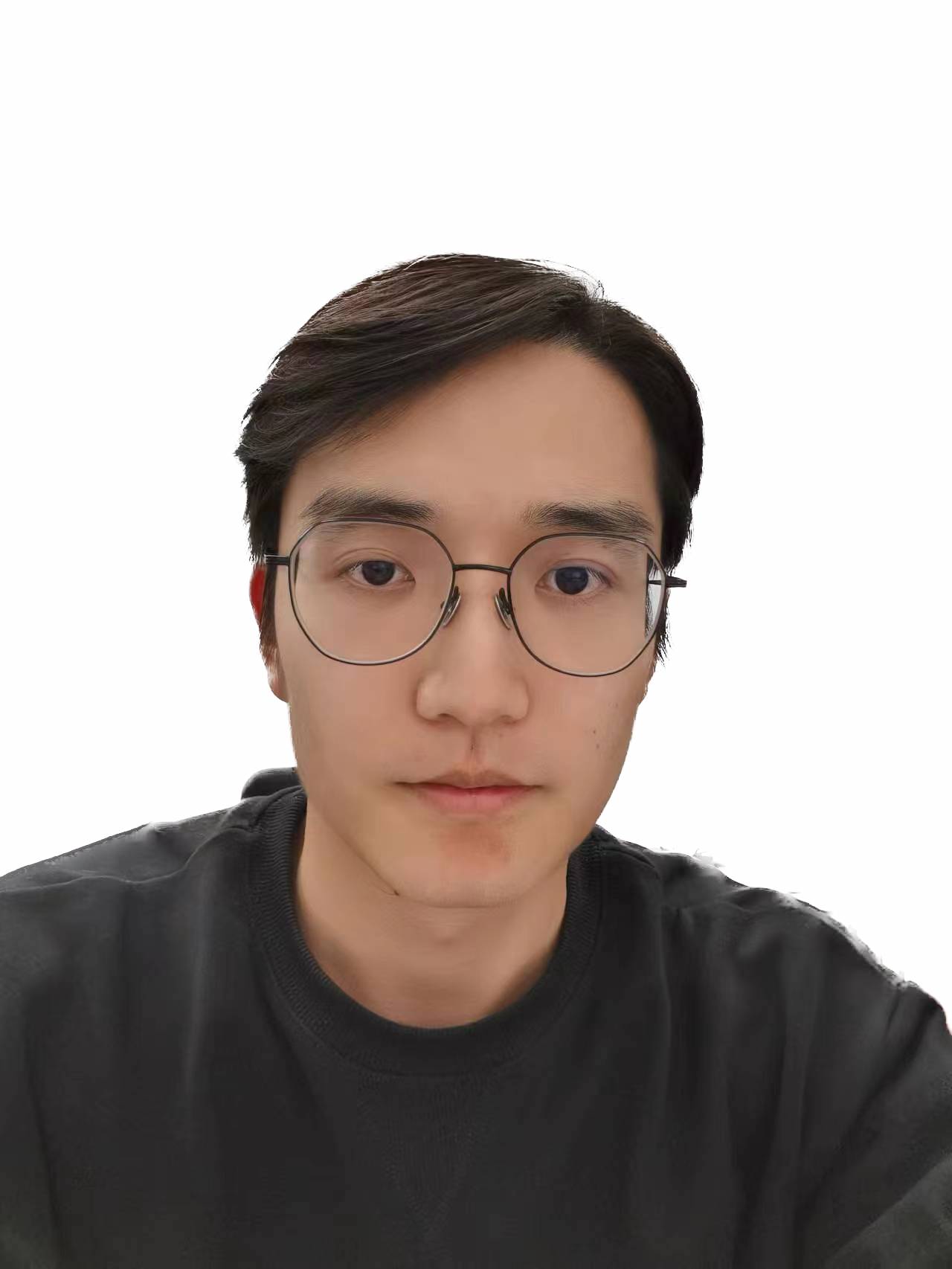}}]{Wang Zhao}
is a Ph.D. student with the Department of Computer Science and Technology, Tsinghua University, China. He received his BEng degree from the Electronic Engineering Department, Tsinghua University, China in 2019. His research interests include 3D computer vision.\end{IEEEbiography}

\vfill\eject

\begin{IEEEbiography}
[{\includegraphics[width=1in,height=1.25in,clip,keepaspectratio]{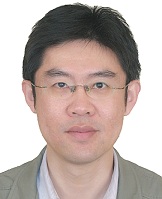}}]
{Yong-Jin Liu} is a Professor with the Department of Computer Science and Technology, Tsinghua University, China. He received the BEng degree from Tianjin University, China, in 1998, and the PhD degree from the Hong Kong University of Science and Technology, Hong Kong, China, in 2004. His research interests include affective computing, computer graphics and computer vision. He is a senior member of IEEE and ACM. For more information, visit
\url{https://cg.cs.tsinghua.edu.cn/people/~Yongjin/Yongjin.html}
\end{IEEEbiography}

\begin{IEEEbiography}[{\includegraphics[width=1in,height=1.25in,clip,keepaspectratio]{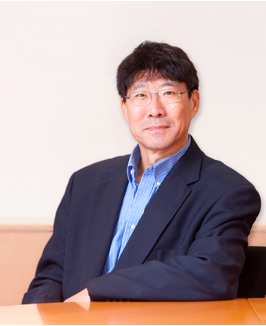}}]
{Wenping Wang} (Fellow, IEEE) received the Ph.D. degree in computer science from the University of Alberta in 1992. He is a Professor of computer science at Texas A\&M University. His research interests include computer graphics, computer visualization, computer vision, robotics, medical image processing, and geometric computing. He is or has been an journal associate editor of ACM Transactions on Graphics, IEEE Transactions on Visualization and Computer Graphics, Computer Aided Geometric Design, and Computer Graphics Forum (CGF). He has chaired a number of international conferences, including Pacific Graphics, ACM Symposium on Physical and Solid Modeling (SPM), SIGGRAPH and SIGGRAPH Asia. Prof. Wang received the John Gregory Memorial Award for his contributions to geometric modeling. He is an IEEE Fellow and an ACM Fellow.
\end{IEEEbiography}

\vfill

\end{document}